\documentclass{article}

     \usepackage[final]{neurips_2021}

\usepackage[utf8]{inputenc} 
\usepackage[T1]{fontenc}    
\usepackage{hyperref}       
\usepackage{url}            
\usepackage{booktabs}       
\usepackage{amsfonts}       
\usepackage{nicefrac}       
\usepackage{microtype}      
\usepackage{xcolor}         
\usepackage{graphicx}
\usepackage{subfigure}
\usepackage{wrapfig}
\usepackage{multirow}
\usepackage{amsmath}
\usepackage{cases}
\usepackage{floatrow}
\newfloatcommand{capbtabbox}{table}[][\FBwidth]

\title{CentripetalText: An Efficient Text Instance Representation for Scene Text Detection}

\author{%
  Tao Sheng, Jie Chen, Zhouhui Lian\thanks{Corresponding author} \\
  Wangxuan Institute of Computer Technology \\
  Peking University, Beijing, China \\
  \texttt{\{shengtao, jiechen01, lianzhouhui\}@pku.edu.cn}
}

\begin{document}

\maketitle

\begin{abstract}
	Scene text detection remains a grand challenge due to the variation
	in text curvatures, orientations, and aspect ratios. One of the hardest
	problems in this task is how to represent text instances of arbitrary
	shapes. Although many methods have been proposed to
	model irregular texts in a flexible manner, most of them lose
	simplicity and robustness. Their complicated post-processings and
	the regression under Dirac delta distribution undermine the detection
	performance and the generalization ability. In this paper, we propose
	an efficient text instance representation named CentripetalText (CT),
	which decomposes text instances into the combination of text kernels
	and centripetal shifts. Specifically, we utilize the centripetal shifts
	to implement pixel aggregation, guiding the external text pixels
	to the internal text kernels. The relaxation operation is
	integrated into the dense regression for centripetal shifts, allowing
	the correct prediction in a range instead of a specific value. The convenient
	reconstruction of text contours and the tolerance of prediction
	errors in our method guarantee the high detection accuracy and the fast
	inference speed, respectively. Besides, we shrink our text detector
	into a proposal generation module, namely CentripetalText Proposal
	Network (CPN), replacing Segmentation Proposal Network (SPN) in Mask TextSpotter v3 and producing
	more accurate proposals. To validate the effectiveness of our method,
	we conduct experiments on several commonly used scene text benchmarks, including
	both curved and multi-oriented text datasets. For the task of scene text
	detection, our approach achieves superior or competitive
	performance compared to other existing methods, e.g., F-measure of
	86.3\% at 40.0 FPS on Total-Text, F-measure of 86.1\% at 34.8 FPS
	on MSRA-TD500, etc. For the task of end-to-end scene text recognition,
	our method outperforms Mask TextSpotter v3 by 1.1\% in F-measure on Total-Text.
\end{abstract}


\section{Introduction}

In the past decade, scene text detection has attracted increasing interests
in the computer vision community, as localizing the region of each text instance
in natural images with high accuracy is an essential prerequisite for many
practical applications such as blind navigation, scene understanding, and
text retrieval. With the rapid development of object
detection~\cite{fasterrcnn,ssd,maskrcnn,fpn} and
segmentation~\cite{fcn,bisenet,dice,zhaoetal}, many promising
methods~\cite{east,textsnake,psenet,DB,PAN,contournet} have been proposed
to solve the problem. However, scene text detection is still a
challenging task due to the variety of text curvatures, orientations, and
aspect ratios.

How to represent text instances in real imagery is one of the major
challenges for scene text detection, and usually there are two strategies
to solve the problem arising from this challenge. The first is to
treat text instances as a specific kind of object and use rotated rectangles
or quadrangles for description. This kind of methods are typically inherited
from generic object detection and often utilize manually designed anchors for
better regression. Obviously, this solution ignores the geometric traits of
irregular texts, which may introduce considerable background noises, and
furthermore, it is difficult to formulate appropriate anchors to fit the
texts of various shapes. The other strategy is to decompose text instances into
several conceptual or physical components, and reconstruct the polygonal
contours through a series of indispensable post-processing steps. For example, PAN~\cite{PAN}
follows the idea of clustering and aggregates text pixels according to
the distances between their embeddings. In TextSnake~\cite{textsnake}, text
instances are represented by text center lines and ordered disks. Consequently,
these methods are more flexible and more general than the previous ones
in modeling. Nevertheless, most of them
suffer from slow inference speed, due to complicated post-processing
steps, essentially caused by this kind of tedious multi-component-based
representation strategy. For another, their component prediction is modeled as a simple Dirac
delta distribution, which strictly requires numerical outputs to reach the exact
positions and thus weakens the ability to tolerate mistakes. The wrong component
prediction will propagate errors to heuristic post-processing procedures,
making the rebuilt text contours inaccurate. Based on the above
observations, we can find out that the implementation of a fast and accurate scene
text detector heavily depends on a simple but effective text instance
representation with a robust post-processing algorithm, which can tolerate
ambiguity and uncertainty.

\begin{wrapfigure}{l}{7cm}
  \centering
  \includegraphics[width=7cm]{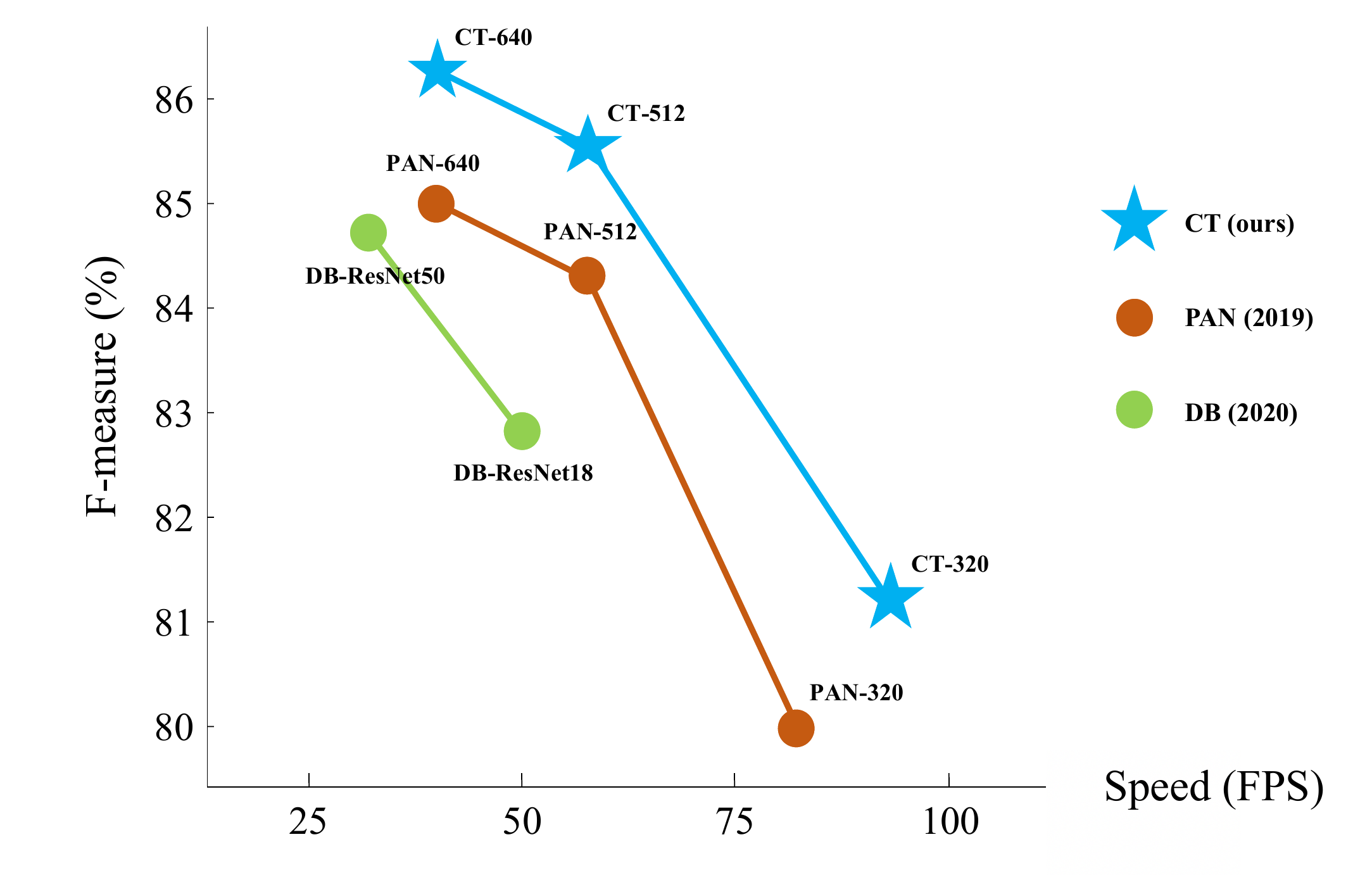}
  \caption{Performance and speed of some top-performing real-time scene
  text detectors on the Total-Text dataset. Enabled by the proposed CT,
  an efficient representation of text instances, our method outperforms
  DB~\cite{DB} and PAN~\cite{PAN}, and achieves the best trade-off
  between accuracy and speed. More results are shown in Tab.~\ref{tab:ttctw}.}
\label{fig:rank}
\end{wrapfigure}

To overcome these problems, we propose an efficient component-based representation
method named CentripetalText (CT) for arbitrary-shaped texts. Enabled by the proposed CT,
our scene text detector outperforms other state-of-the-art approaches and achieves
the best trade-off between accuracy and speed (as shown in Fig.~\ref{fig:rank}).
Specifically, as illustrated in Fig.~\ref{fig:pipeline}, our method consists of two steps:
i) input images are fed to the convolutional neural network to predict the probability
maps and centripetal shift maps. ii) pixels are grouped to form the text
instances through the heuristics based on text kernels and centripetal shifts.
In details, the text kernels are generated from the probability map followed by
binarization and connected component search, and the centripetal shifts are predicted
at each position of the centripetal shift map. Then each pixel is shifted by the
amount of its centripetal shift from its original position in the centripetal
shift map to the text kernel pixel or background pixel in the probability map.
All pixels that can be shifted into the region of the same text kernel form a
text instance. In this manner, we can reconstruct the final text instances fast and
easily through marginal matrix operations and several calls to functions of the
OpenCV library. Moreover, we develop an enhanced regression
loss, namely the Relaxed L1 Loss, mainly for dense centripetal shift regression,
which further improves the detection precision. Benefiting from the new loss, our
method is robust to the prediction errors of centripetal shifts because the
centripetal shifts which can guide pixels to the region of the right text kernel
are all regarded as positive. Besides, CT can be fused with
CNN-based text detectors or spotters in a plug-and-play manner. We replace SPN in Mask
TextSpotter v3~\cite{masktextspotterv3} with our CentripetalText Proposal Network (CPN),
a proposal generation module based on CT, which produces more accurate proposals and
improves the end-to-end text recognition performance further.

To evaluate the effectiveness of the proposed CT and Relaxed L1 Loss,
we adopt the design of network architecture in PAN~\cite{PAN} and train a powerful
end-to-end scene text detector by replacing its text instance representation and loss
function with ours. We conduct extensive experiments on the commonly used
scene text benchmarks including Total-Text~\cite{totaltext}, CTW1500~\cite{ctw1500},
and MSRA-TD500~\cite{msratd500}, demonstrating that our method achieves
superior or competitive performance compared to the state of the art, e.g., F-measure of 86.3\%
at 40.0 FPS on Total-Text, F-measure of 86.1\% at 34.8 FPS on MSRA-TD500, etc.
For the task of end-to-end text recognition, equipped with CPN, the F-measure value
of Mask TextSpotter v3 can further be boosted to 71.9\% and 79.5\% without
and with lexicon, respectively.

Major contributions of our work can be summarized as follows:
\begin{itemize}
	\item We propose a novel and efficient text instance representation
	method named CentripetalText, in which text instances are decomposed into
	the combination of text kernels and centripetal shifts. The attached
	post-processing algorithm is also simple and robust, making the generation
	of text contours fast and accurate.
	\item To reduce the burden of model training, we develop an
	enhanced loss function, namely the Relaxed L1 Loss, mainly for dense
	centripetal shift regression, which further improves the detection performance.
	\item Equipped with the proposed CT and the Relaxed L1 Loss,
	our scene text detector achieves superior or competitive results compared to other
	existing approaches on the curved or oriented text benchmarks, and our
	end-to-end scene text recognizer surpasses the current state of the art.
\end{itemize}


\section{Related work}

Text instance representation methods can be roughly
classified into two categories: component-free methods and component-based methods.

\textbf{Component-free methods} treat every text instance as a complete
whole and directly regress the rotated rectangles or quadrangles for describing
scene texts without any reconstruction process. These methods are usually
inspired by general object detectors such as Faster R-CNN~\cite{fasterrcnn} and
SSD~\cite{ssd}, and often utilize heuristic anchors as prior knowledge.
TextBoxes~\cite{textboxes} successfully adapted the object detection framework
SSD for text detection by modifying the aspect ratios of anchors and the kernel
scales of filters. TextBoxes++~\cite{textboxes++} and EAST~\cite{east} could
predict either rotated rectangles or quadrangles for text regions with and
without the prior knowledge of anchors, respectively. SPCnet~\cite{spcnet} modified
Mask R-CNN~\cite{maskrcnn} by adding the semantic segmentation guidance to
suppress false positives.

\textbf{Component-based methods} prefer to model text instances from local
perspectives and decompose instances into components such as characters or
text fragments. SegLink~\cite{seglink} decomposed long texts into locally-detectable
segments and links, and combined the segments into whole words according to
the links to get final detection results. MSR~\cite{MSR} detected scene texts by
predicting dense text boundary points. BR~\cite{BR} further improved MSR by 
regressing the positions of boundary points from two opposite directions.
PSENet~\cite{psenet} gradually expanded the detected areas from small kernels 
to complete instances via a progressive scale expansion algorithm.


\begin{figure*}[t]
  \centering
  \includegraphics[width=\linewidth]{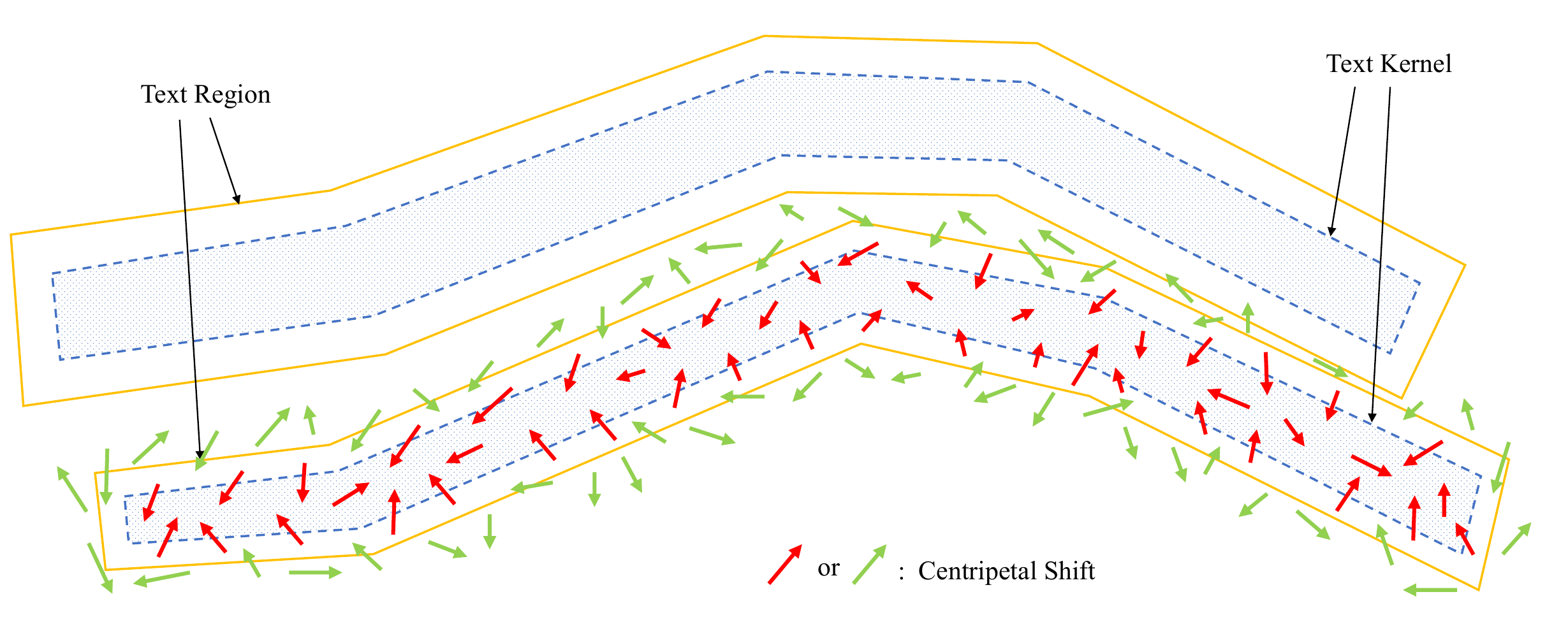}
  \caption{Illustration of the proposed CentripetalText representation. Text
  regions (in yellow) can be decomposed into the combination of text kernels
  (in blue) and centripetal shifts (either in red or green). The centripetal
  shifts represented as green arrows start from background pixels to non-text-kernel
  pixels, which are useless for the further generation of text contours, while
  other centripetal shifts in red start from text region
  (foreground) pixels to text kernel pixels, which contribute to define the
  shapes. All the pixels that can be shifted into the region of the same text
  kernel form a text instance. For the sake of better demonstration, we only visualize the
  centripetal shifts over the bottom text instance.}
\label{fig:represent}
\end{figure*}

\section{Methodology}

In this section, we first introduce our new representation (i.e., CT)
for texts of arbitrary shapes. Then, we elaborate on our method and training
details.

\subsection{Representation}

\paragraph{Overview}

An efficient scene text detector must have a well-defined representation for
text instances. The traditional description methods inherited from generic
object detection (e.g.,
rotated rectangles or quadrangles) fail to encode the geometric properties of
irregular texts. To guarantee the flexibility and generality, we propose a new
method named CentripetalText, in which text instances are composed of text
kernels and centripetal shifts. As demonstrated in Fig.~\ref{fig:represent},
our CT expresses a text instance as a cluster of the pixels which
can be shifted into the region of the same text kernel through centripetal
shifts. As pixels are the basic units of digital images, CT
has the ability to model different forms of text instances, regardless of their shapes and lengths.

Mathematically, given an input image $I$, the ground-truth annotations are
denoted as $\{T_1,T_2,..,T_i,..\}$, where $T_i$ stands for the $i$th text
instances. Each text instance $T_i$ has its corresponding text kernel $K_i$,
a shrunk version of the original text region. Since a text kernel is
a subset of its text instance, which satisfies $K_i \subseteq T_i$, we treat
it as the main basic of the pixel aggregation. Different against the distance in conventional
methods, each centripetal shift $s_j$ which appears in a position $p_j$
of the image guides the clustering of text pixels. In this sense, the text instance
$T_i$ can be easily represented with the aggregated pixels which can be shifted
into the region of a text kernel according to the values of their centripetal shifts:
\begin{equation}
	T_i = \{p_j \mid (p_j+s_j)\in K_i\}.
\end{equation}

\paragraph{Label generation}
\label{lg}

\begin{figure*}[t]
	\centering
	\subfigure[]{
		\includegraphics[width=0.21\linewidth]{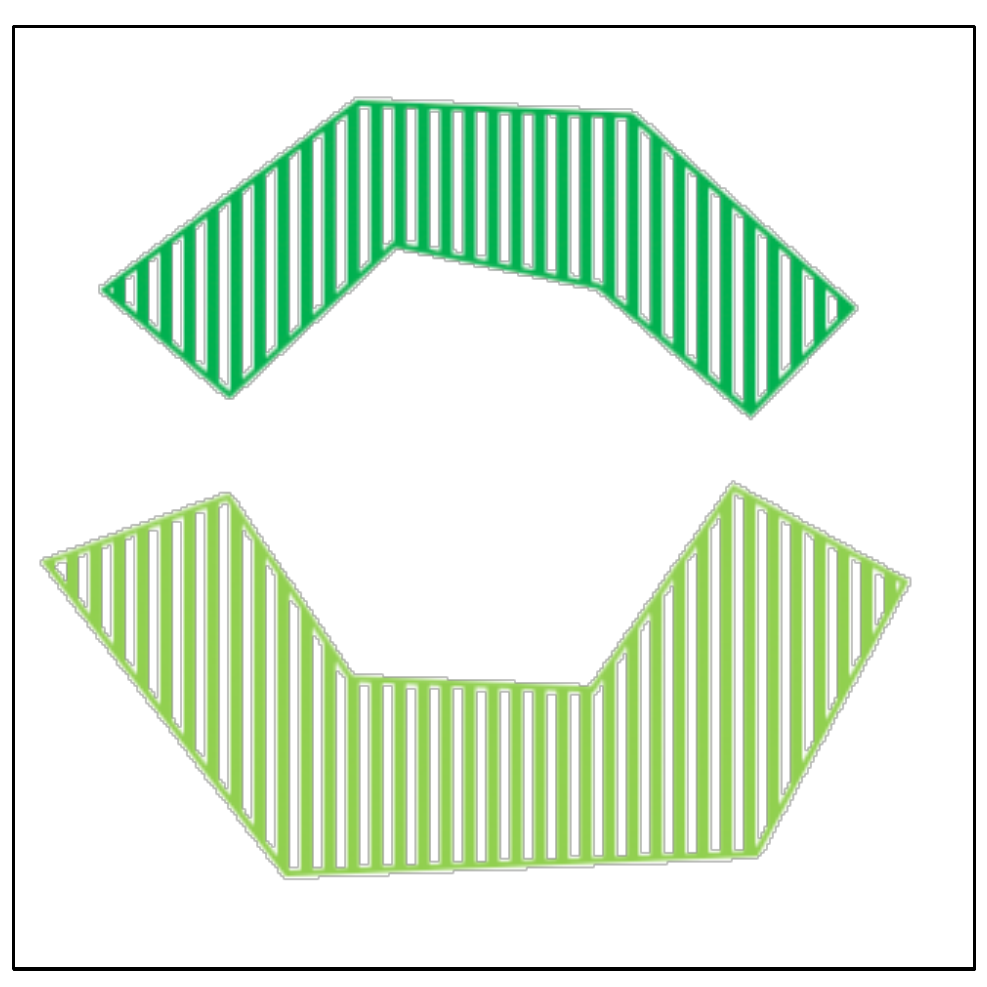}
		\label{fig:labelgenea}
	}
	\quad
	\subfigure[]{
		\includegraphics[width=0.21\linewidth]{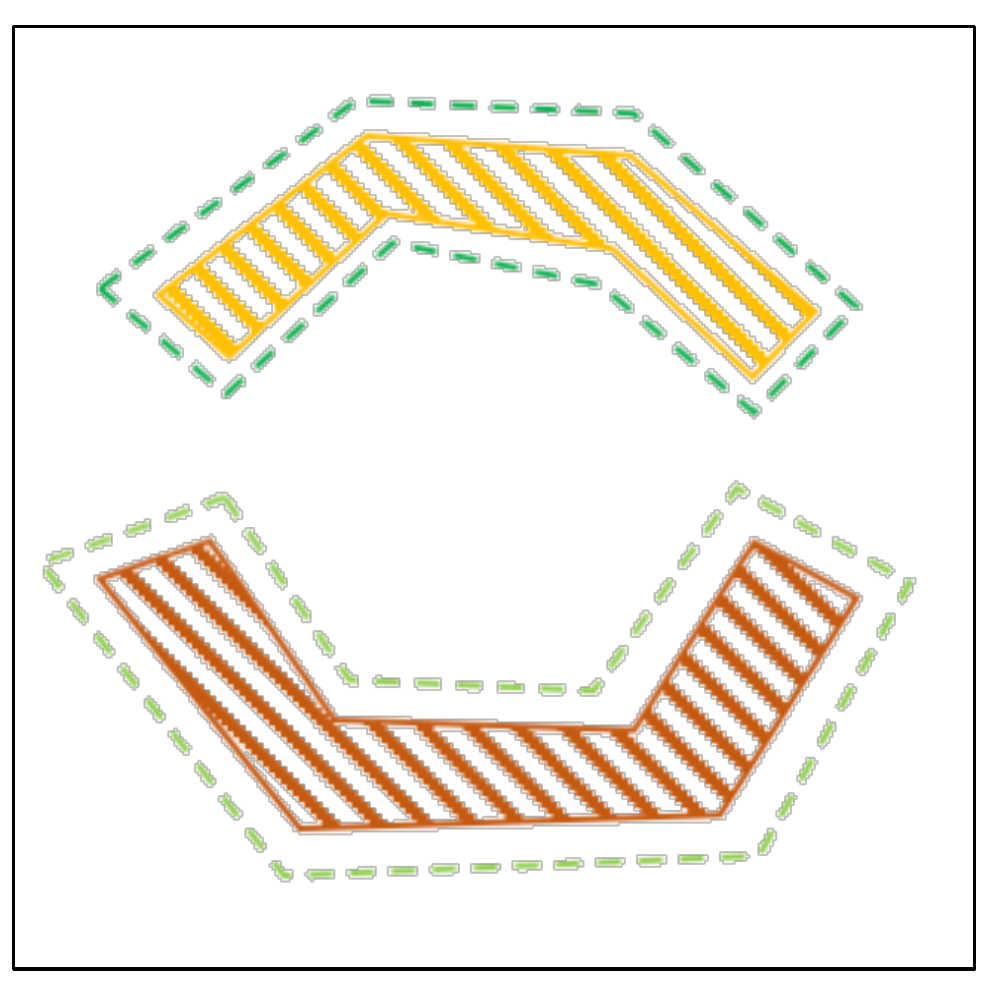}
		\label{fig:labelgeneb}
	}
	\quad
	\subfigure[]{
		\includegraphics[width=0.21\linewidth]{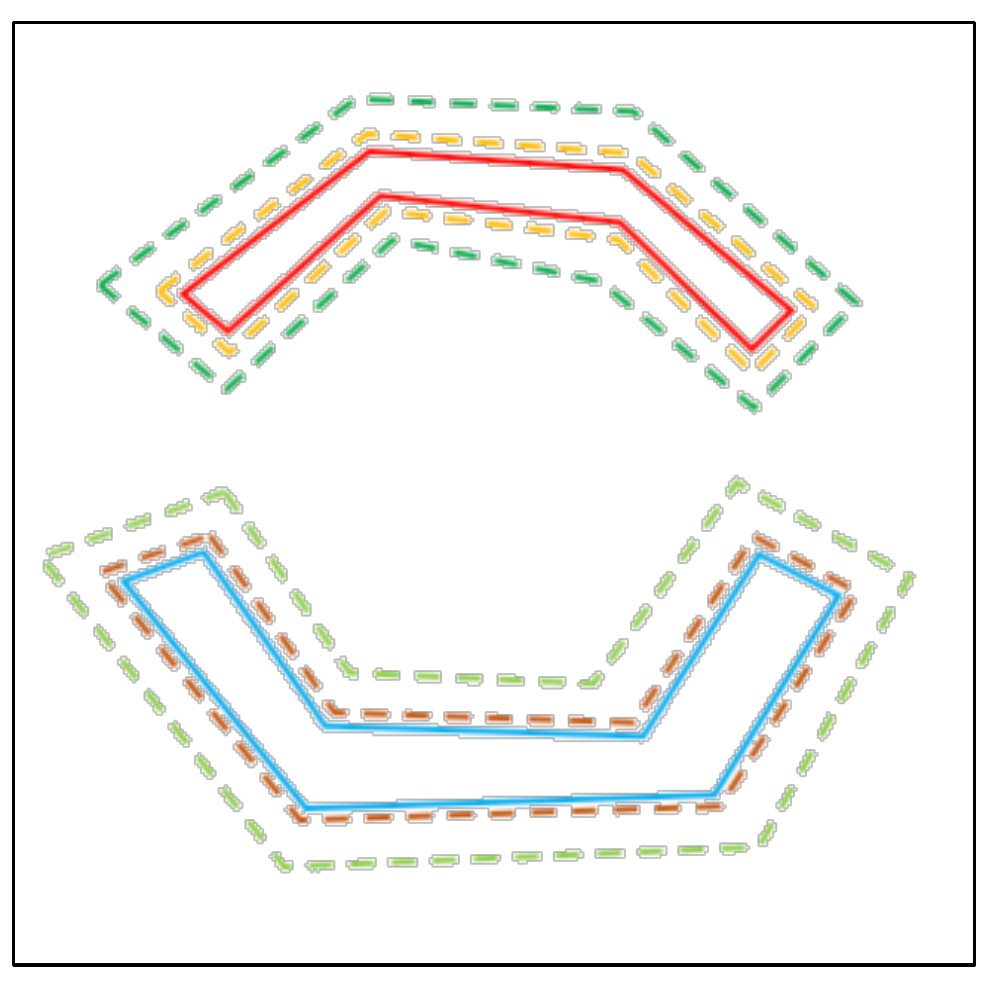}
		\label{fig:labelgenec}
	}
	\quad
	\subfigure[]{
		\includegraphics[width=0.21\linewidth]{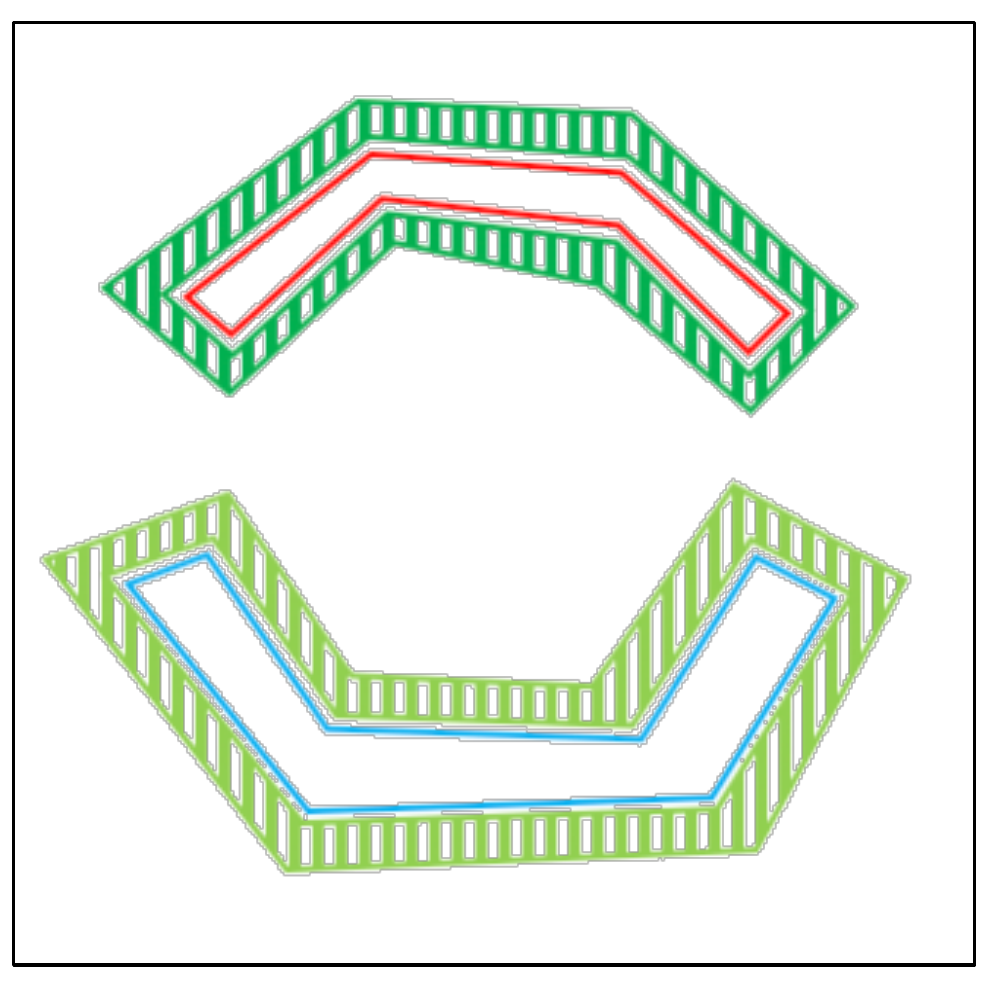}
		\label{fig:labelgened}
	}
	\caption{Label generation. (a) Text instance; (b) Text kernel; (c) Text
	kernel reference; (d) Text region that contains nonzero centripetal shifts.}
	\label{fig:labelgene}
\end{figure*}

The label generation for the probability map is inspired by PSENet~\cite{psenet},
where the positive area of the text kernel (shaded areas in Fig.~\ref{fig:labelgeneb})
is generated by shrinking the annotated polygon (shaded areas in Fig.~\ref{fig:labelgenea})
using the Vatti clipping algorithm~\cite{vatti}. The offset of shrinking is
computed based on the perimeter and area of the original polygon and the
shrinking ratio is set to 0.7 empirically. Since the annotations in the dataset
may not perfectly fit the text instances well, we design a training mask
$M$ to distinguish the supervision of the valid and ignoring regions.
The text instance excluding the text kernel ($T_i-K_i$) is the ignoring region,
which means that the gradients in this area are not propagated back to the network.
The training mask $M$ can be formulated as follows:
\begin{equation}
M_j =
\begin{cases}
	0, & \text{if } p_j \in \bigcup\limits_i(T_i-K_i) \\
	1, & \text{otherwise}.
\end{cases}
\end{equation}
We simply multiply the training mask by the loss of the segmentation branch to
eliminate the influence brought by wrong annotations.

In the label generation step of the regression branch, the text instance affects the centripetal
shift map in three ways. First, the centripetal shift in the background region
($I - \bigcup_i T_i$) should prevent the background pixels from entering into
any text kernel and thus we set it to $(0, 0)$ intuitively. Second, the centripetal shift
in the text kernel region ($\bigcup_i K_i$) should keep the locations of text kernel pixels
unchanged and we also set it to $(0, 0)$ for convenience. Third, we expect
that each pixel in the region of the text instance excluding the text kernel
($\bigcup_i(T_i-K_i)$) can be guided to its corresponding kernel by the centripetal
shift. Therefore, we continuously conduct the erosion operation over the text
kernel map twice, compare these two temporary results and obtain the text kernel
reference (polygons with solid lines in Fig.~\ref{fig:labelgenec}) as the destination of the
generated centripetal shift. As shown in Fig.~\ref{fig:labelgened}, we build the
centripetal shift between each pixel in the shaded area and its nearest text kernel
reference to prevent numerical accuracy issues caused by rounding off. Note
that if two instances overlap, the smaller instance has the higher priority.
Specifically, the centripetal shift $s_j$ can be calculated by:
\begin{equation}
s_j =
\begin{cases}
	(0, 0), & \text{if } p_j \in (I - \bigcup\limits_i T_i)\\
	\overrightarrow{p_jp_j^*}, & \text{if } p_j \in \bigcup\limits_i(T_i-K_i) \\
	(0, 0), & \text{if } p_j \in \bigcup\limits_i K_i ,
\end{cases}
\end{equation}
where $p_j^*$ represents the nearest text kernel reference to the pixel $p_j$.
During training, the Smooth L1 loss~\cite{fastrcnn} is applied for supervision.
Nevertheless, according to a previous observation~\cite{GFL}, the dense regression
can be modeled as a simple Dirac delta distribution, which fails to consider
the ambiguity and uncertainty in the training data. To address the problem, we develop a
regression mask $R$ for the relaxation operation and integrate it into the Smooth
L1 loss to reduce the burden of model training. We extend the correct prediction
from one specific value to a range and any centripetal shift which moves the pixel
into the right region is treated as positive during training. The regression mask
can be formulated as follows:
\begin{equation}
R_j =
\begin{cases}
	0, & \text{if } p_j \in T_i \text{ and } (p_j + \widehat{s_j}) \in K_i \\
	   & \text{or } p_j \not\in \bigcup\limits_i T_i \text{ and } (p_j + \widehat{s_j}) \not\in \bigcup\limits_i K_i \\
	1, & \text{otherwise}.
\end{cases}
\end{equation}
where $\widehat{s_j}$ and $s_j$ denote the predicted centripetal shift at the
position $j$ and its ground truth, respectively.
Like the segmentation loss, we multiply the regression mask by the Smooth L1 loss and
form a novel loss function, namely the Relaxed L1 loss for dense centripetal shift
prediction, to further improve the detection accuracy. The Relaxed L1 loss
function can be formulated as follows:
\begin{equation}
\mathcal{L}_{regression} = \sum\limits_j \big( R_j \cdot \text{Smooth}_\text{L1}(s_j, \widehat{s_j})\big),
\end{equation}
where $\text{Smooth}_\text{L1}()$ denotes the standard Smooth L1 loss.

\begin{figure*}[t]
  \centering
  \includegraphics[width=0.8\linewidth]{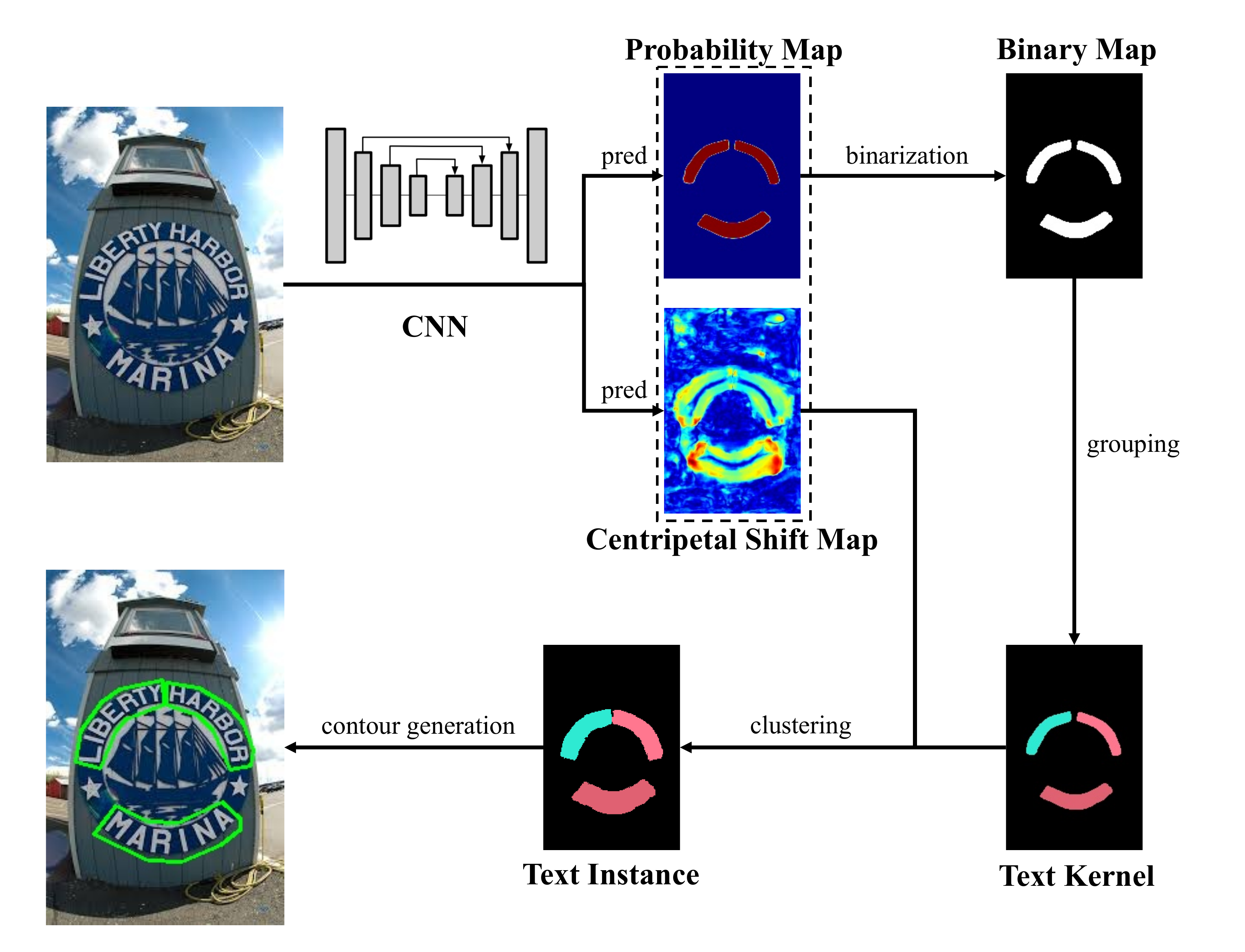}
  \caption{An overview of our proposed model.}
\label{fig:pipeline}
\end{figure*}

\subsection{Scene text detection with CentripetalText}

\paragraph{Network Architecture}

In order to detect texts with arbitrary shapes fast and accurately, we adopt the
efficient model design in PAN~\cite{PAN} and equip it with our CT
and Relaxed L1 loss. First, ResNet18~\cite{resnet} is used as the default backbone
for fair comparison. Then, to remedy the weak representation ability of the
lightweight backbone, two cascaded FPEMs~\cite{PAN} are utilized to continuously enhance the
feature pyramid in both top-down and bottom-up manners. Afterwards, the generated
feature pyramids of different depths are fused by FFM~\cite{PAN} into a single
basic feature. Finally, we predict the probability map
and the centripetal shift map from the basic feature for further contour generation.

\paragraph{Inference}

The procedure of text contour reconstruction is shown in Fig.~\ref{fig:pipeline}.
After feed-forwarding, the network produces the probability map and the centripetal
shift map. We firstly binarize the probability map with a constant threshold
(0.2) to get the binary map. Then, we find the connected components (text
kernels) from the binary map as the clusters of pixel aggregation. Afterwards,
we assign each pixel to the corresponding cluster according to which text kernel (or background)
can the pixel be shifted into by its centripetal shift. Finally, we build the
text contour for each group of text pixels. Note that our post-processing strategy
has an essential difference against PAN~\cite{PAN}. The post-processing in PAN is
an iterative process, which gradually extends the text kernel to the text region
by merging its neighbor pixels iteratively. On the contrary, we conduct the aggregation
in one step, which means that the centripetal shifts of all pixels can be calculated in parallel by implementing
one matrix operation, saving the inference time to a certain extent.

\paragraph{Optimization}

Our loss function can be formulated as:
\begin{equation}
\mathcal{L} = \mathcal{L}_{segmentation} + \lambda\mathcal{L}_{regression},
\end{equation}
where $\mathcal{L}_{segmentation}$ denotes the segmentation loss of text
kernels, and $\mathcal{L}_{regression}$ denotes the regression loss of centripetal
shifts. $\lambda$ is a constant to balance the weights of the
segmentation and regression losses. We set it to 0.05 in all experiments. Specifically,
the prediction of text kernels is basically a pixel-wise binary classification
problem and we apply the dice loss~\cite{dice} to handle this task. Equipped with
the training mask $M$, the segmentation loss can be defined as:
\begin{equation}
\mathcal{L}_{segmentation}=\sum_j \big(M_j \cdot \text{Dice}(c_j, \widehat{c_j})\big),
\end{equation}
where $\text{Dice}()$ denotes the dice loss function, $\widehat{c_j}$ and $c_j$
denote the predicted probability of text kernels at the position $j$ and its
ground truth, respectively. Note that we adopt Online Hard Example Mining
(OHEM)~\cite{ohem} to address the imbalance of positives and negatives while calculating
$\mathcal{L}_{segmentation}$. Regarding the regression loss, a detailed
description has been provided in Sec.~\ref{lg}.

\paragraph{CentripetalText Proposal Network}

Our scene text detector is shrunk to a text proposal
module, termed as CentripetalText Proposal Network (CPN), by transforming the
polygonal outputs to the minimum area rectangles and instance masks. We follow
the main design of the text detection and recognition modules of Mask TextSpotter
v3~\cite{masktextspotterv3} and replace SPN with our CPN for the comparison of
proposal quality and recognition accuracy.


\section{Experiments}

\subsection{Datasets}
\label{sec:data}

\textbf{SynthText}~\cite{synthtext} is a synthetic dataset, consisting of more
than 800,000 synthetic images. This dataset is used to pre-train our model.

\textbf{Total-Text}~\cite{totaltext} is a curved text dataset including 1,255
training images and 300 testing images. This dataset contains horizontal,
multi-oriented, and curve text instances labeled at the word level.

\textbf{CTW1500}~\cite{ctw1500} is another curved text dataset including 1,000
training images and 500 testing images. The text instances are annotated at
text-line level with 14-polygons.

\textbf{MSRA-TD500}~\cite{msratd500} is a multi-oriented text dataset which
contains 300 training images and 200 testing images with text-line level annotation.
Due to its small scale, we follow the previous works~\cite{east,textsnake} to
include 400 extra training images from HUST-TR400~\cite{hust}.

\subsection{Implementation details}
\label{sec:id}

To make fair comparisons, we use the same training settings described below.
ResNet~\cite{resnet} pre-trained on ImageNet~\cite{imagenet} is used as the
backbone of our method. All models are optimized by the Adam optimizer
with the batch size of 16 on 4 GPUs. We train our model under two training
strategies: (1) learning from scratch; (2) fine-tuning models pre-trained on the
SynthText dataset. Whichever training strategies, we pre-train models on SynthText
for 50k iterations with a fixed learning rate of $1\times{10}^{-3}$, and
train models on real datasets for 36k iterations with the “poly” learning
rate strategy~\cite{zhaoetal}, where “power” is set to 0.9 and the initial
learning rate is $1\times{10}^{-3}$. We follow the official
implementation of PAN to implement data augmentation, including random scale, random horizontal flip,
random rotation, and random crop. The blurred texts labeled as DO NOT CARE
are ignored during training. In addition, we set the negative-positive
ratio of OHEM to 3, and the shrinking rate of text kernel to 0.7.
All those models are tested with a batch size of 1 on a GTX 1080Ti GPU without
bells and whistles. For the task of end-to-end scene text recognition, we leave the original training
and testing settings of Mask TextSpotter v3 unchanged.

\subsection{Ablation study}

To analyze our designs in depth, we conduct a series of ablation
studies on both curve and multi-oriented text datasets (Total-Text and
MSRA-TD500). In addition, all models in this subsection are trained
from scratch.

\begin{table*}[h]
	\begin{floatrow}
	\capbtabbox{
		\resizebox{0.43\textwidth}{!}{
 		\begin{tabular}{c|c|cc|cc}
 			\hline
			\multirow{2}{*}{Backbone} & \multirow{2}{*}{Neck} & \multicolumn{2}{c|}{Total-Text} & \multicolumn{2}{c}{MSRA-TD500} \\
			\cline{3-6}
			 & & F & FPS & F & FPS \\
			\hline
			\hline
			\multirow{2}{*}{ResNet18} & FPN~\cite{fpn}  & 84.0 & \textbf{46.8} & 81.6 & \textbf{42.0} \\
			 & FPEM~\cite{PAN} & 84.9 & 40.0 & 83.0 & 34.8 \\
			\hline
			\multirow{2}{*}{ResNet50} & FPN~\cite{fpn}  & 85.1 & 25.5 & 82.7 & 21.9 \\
			 & FPEM~\cite{PAN} & \textbf{85.6} & 24.0 & \textbf{83.5} & 20.5 \\
 			\hline
 		\end{tabular}
 		}
	}{
 		\caption{Quantitative results of our text detection models with different backbones
 		and necks. “F” means F-measure.}
 		\label{tab:structure}
	}
	\capbtabbox{
		\resizebox{0.47\textwidth}{!}{
 		\begin{tabular}{l|l|cc|cc}
 			\hline
 			\multirow{2}{*}{Rep.} & \multirow{2}{*}{Regression Loss} & \multicolumn{2}{c|}{Total-Text} & \multicolumn{2}{c}{MSRA-TD500} \\
 			\cline{3-6}
			 & & F & FPS & F & FPS \\
			\hline
			\hline
			PAN~\cite{PAN} & & 83.5 & 39.6 & 78.9 & 30.2 \\
			\hline
			\multirow{3}{*}{CT (Ours)} & Smooth L1\cite{fastrcnn} & 83.0 & \textbf{40.0} & 81.0 & \textbf{34.8} \\
			 & Balanced L1\cite{librarcnn} & 83.8 & \textbf{40.0} & 81.7 & \textbf{34.8} \\
			 & Relaxed L1 & \textbf{84.9} & \textbf{40.0} & \textbf{83.0} & \textbf{34.8} \\
 			\hline
		\end{tabular}
		}
	}{
 		\caption{Comparison between PAN~\cite{PAN} and our models with
 		different regression losses. “Rep.” denotes representation
 		and “F” means F-measure.}
 		\label{tab:loss}
	}
	\end{floatrow}
\end{table*}

On the one hand, to make full use of the capability of the proposed CT,
we try different backbones and necks to find the best network architecture.
As shown in Tab.~\ref{tab:structure}, although “ResNet18 + FPN” and
“ResNet50 + FPEM” are the fastest and most accurate detectors,
respectively, “ResNet18 + FPEM” achieves the best trade-off between
accuracy and speed. Thus, we keep this combination by default in the
following experiments. On the other hand, we study the validity of the Relaxed
L1 loss by replacing it with others. Compared with the baseline Smooth
L1 Loss~\cite{fastrcnn} and the newly-released Balanced L1 loss~\cite{librarcnn},
the F-measure value of our method improves over 1\% on both two datasets, which
indicates the effectiveness of the Relaxed L1 loss. Moreover, under the
same settings of the model architecture, our method outperforms PAN by a wide
extent while keeping its fast inference speed, indicating that the proposed CT is
more efficient.

\newpage

\subsection{Comparisons with state-of-the-art methods}

\begin{table*}[t]
	\centering
	\caption{Quantitative detection results on Total-Text and CTW1500. “P”, “R” and
	“F” represent the precision, recall, and F-measure, respectively. “Ext.” denotes
	external training data. * indicates the multi-scale testing is performed.}
	\label{tab:ttctw}
	\begin{tabular}{l|c|c|cccc|cccc}
		\hline
		\multirow{2}{*}{Method} & \multirow{2}{*}{Ext.} & \multirow{2}{*}{Venue} & \multicolumn{4}{c|}{Total-Text} & \multicolumn{4}{c}{CTW1500} \\
		\cline{4-11}
		 & & & P & R & F & FPS & P & R & F & FPS \\
		\hline
		CTPN~\cite{CTPN} & - & ECCV’16 & - & - & - & - & 60.4 & 53.8 & 56.9 & 7.1 \\
		SegLink~\cite{seglink} & - & CVPR'17 & 30.3 & 23.8 & 26.7 & - & 42.3 & 40.0 & 40.8 & 10.7 \\
		EAST~\cite{east} & - & CVPR’17 & 50.0 & 36.2 & 42.0 & - & 78.7 & 49.1 & 60.4 & 21.2 \\
		PSENet~\cite{psenet} & - & CVPR’19 & 81.8 & 75.1 & 78.3 & 3.9 & 80.6 & 75.6 & 78.0 & 3.9 \\
		PAN~\cite{PAN} & - & ICCV'19 & 88.0 & 79.4 & 83.5 & 39.6 & 84.6 & 77.7 & 81.0 & 39.8 \\
		\textbf{CT-320} & - & - & 87.6 & 72.7 & 79.4 & \textbf{93.2} & \textbf{85.7} & 73.2 & 79.0 & \textbf{107.2} \\
		\textbf{CT-512} & - & - & 87.9 & 80.8 & 84.2 & 57.0 & 85.2 & 78.4 & 81.7 & 59.8 \\
		\textbf{CT-640} & - & - & \textbf{88.8} & \textbf{81.4} & \textbf{84.9} & 40.0 & 85.5 & \textbf{79.2} & \textbf{82.2} & 40.8 \\
		\hline
		\hline
		TextSnake~\cite{textsnake} & $\checkmark$ & ECCV’18 & 82.7 & 74.5 & 78.4 & - & 67.9 & \textbf{85.3} & 75.6 & - \\
		MSR~\cite{MSR} & $\checkmark$ & IJCAI'19 & 83.8 & 74.8 & 79.0 & - & 85.0 & 78.3 & 81.5 & - \\
		SegLink++~\cite{seglink++} & $\checkmark$ & PR’19 & 82.1 & 80.9 & 81.5 & - & 82.8 & 79.8 & 81.3 & - \\
		PSENet~\cite{psenet} & $\checkmark$ & CVPR'19 & 84.0 & 78.0 & 80.9 & 3.9 & 84.8 & 79.7 & 82.2 & 3.9 \\
		SPCNet~\cite{spcnet} & $\checkmark$ & AAAI'19 & 83.0 & 82.8 & 82.9 & - & - & - & - & - \\
		LOMO*~\cite{LOMO} & $\checkmark$ & CVPR’19 & 87.6 & 79.3 & 83.3 & - & 85.7 & 76.5 & 80.8 & - \\
		CRAFT~\cite{CRAFT} & $\checkmark$ & CVPR’19 & 87.6 & 79.9 & 83.6 & - & 86.0 & 81.1 & 83.5 & - \\
		Boundary~\cite{boundary} & $\checkmark$ & AAAI’20 & 85.2 & 83.5 & 84.3 & - & - & - & - & - \\
		DB~\cite{DB} & $\checkmark$ & AAAI'20 & 87.1 & 82.5 & 84.7 & 32.0 & 86.9 & 80.2 & 83.4 & 22.0 \\
		PAN~\cite{PAN} & $\checkmark$ & ICCV'19 & 89.3 & 81.0 & 85.0 & 39.6 & 86.4 & 81.2 & 83.7 & 39.8 \\
		DRRG~\cite{DRRG} & $\checkmark$ & CVPR’20 & 86.5 & \textbf{84.9} & 85.7 & - & 85.9 & 83.0 & \textbf{84.5} & - \\
		\textbf{CT-320} & $\checkmark$ & - & 88.0 & 75.4 & 81.2 & \textbf{93.2} & 87.7 & 74.7 & 80.7 & \textbf{107.2} \\
		\textbf{CT-512} & $\checkmark$ & - & 90.2 & 81.5 & 85.6 & 57.0 & 87.8 & 79.0 & 83.2 & 59.8 \\
		\textbf{CT-640} & $\checkmark$ & - & \textbf{90.5} & 82.5 & \textbf{86.3} & 40.0 & \textbf{88.3} & 79.9 & 83.9 & 40.8 \\
		\hline
	\end{tabular}
\end{table*}

\begin{wraptable}{r}{7cm}
	\centering
	\caption{Quantitative detection results on MSRA-TD500. “P”, “R” and “F” represent
	the precision, recall, and F-measure, respectively. “Ext.” denotes external training
	data.}
	\label{tab:msra}
\resizebox{7cm}{!}{
	\begin{tabular}{l|c|cccc}
		\hline
		Method & Ext. & P & R & F & FPS \\
		\hline
		RRPN~\cite{rrpn} & - & 82.0 & 68.0 & 74.0 & - \\
		EAST~\cite{east} & - & \textbf{87.3} & 67.4 & 76.1 & 13.2 \\
		PAN~\cite{PAN} & - & 80.7 & 77.3 & 78.9 & 30.2 \\
		\textbf{CT-736} & - & 87.1 & \textbf{79.3} & \textbf{83.0} & \textbf{34.8} \\
		\hline
		\hline
		SegLink~\cite{seglink} & $\checkmark$ & 86.0 & 70.0 & 77.0 & 8.9 \\
		PixelLink~\cite{pixellink} & $\checkmark$ & 83.0 & 73.2 & 77.8 & 3.0 \\
		TextSnake~\cite{textsnake} & $\checkmark$ & 83.2 & 73.9 & 78.3 & 1.1 \\
		RRD~\cite{rrd} & $\checkmark$ & 87.0 & 73.0 & 79.0 & 10.0 \\
		TextField~\cite{textfield} & $\checkmark$ & 87.4 & 75.9 & 81.3 & - \\
		CRAFT~\cite{CRAFT} & $\checkmark$ & 88.2 & 78.2 & 82.9 & 8.6 \\
		MCN~\cite{MCN} & $\checkmark$ & 88.0 & 79.0 & 83.0 & - \\
		PAN~\cite{PAN} & $\checkmark$ & 84.4 & \textbf{83.8} & 84.1 & 30.2 \\
		DB~\cite{DB} & $\checkmark$ & \textbf{91.5} & 79.2 & 84.9 & 32.0 \\
		DRRG~\cite{DRRG} & $\checkmark$ & 88.1 & 82.3 & 85.1 & - \\
		\textbf{CT-736} & $\checkmark$ & 90.0 & 82.5 & \textbf{86.1} & \textbf{34.8} \\
		\hline
	\end{tabular}
}
\end{wraptable}

\paragraph{Curved text detection}

\begin{figure*}[t]
  \centering
  \includegraphics[width=0.78\linewidth]{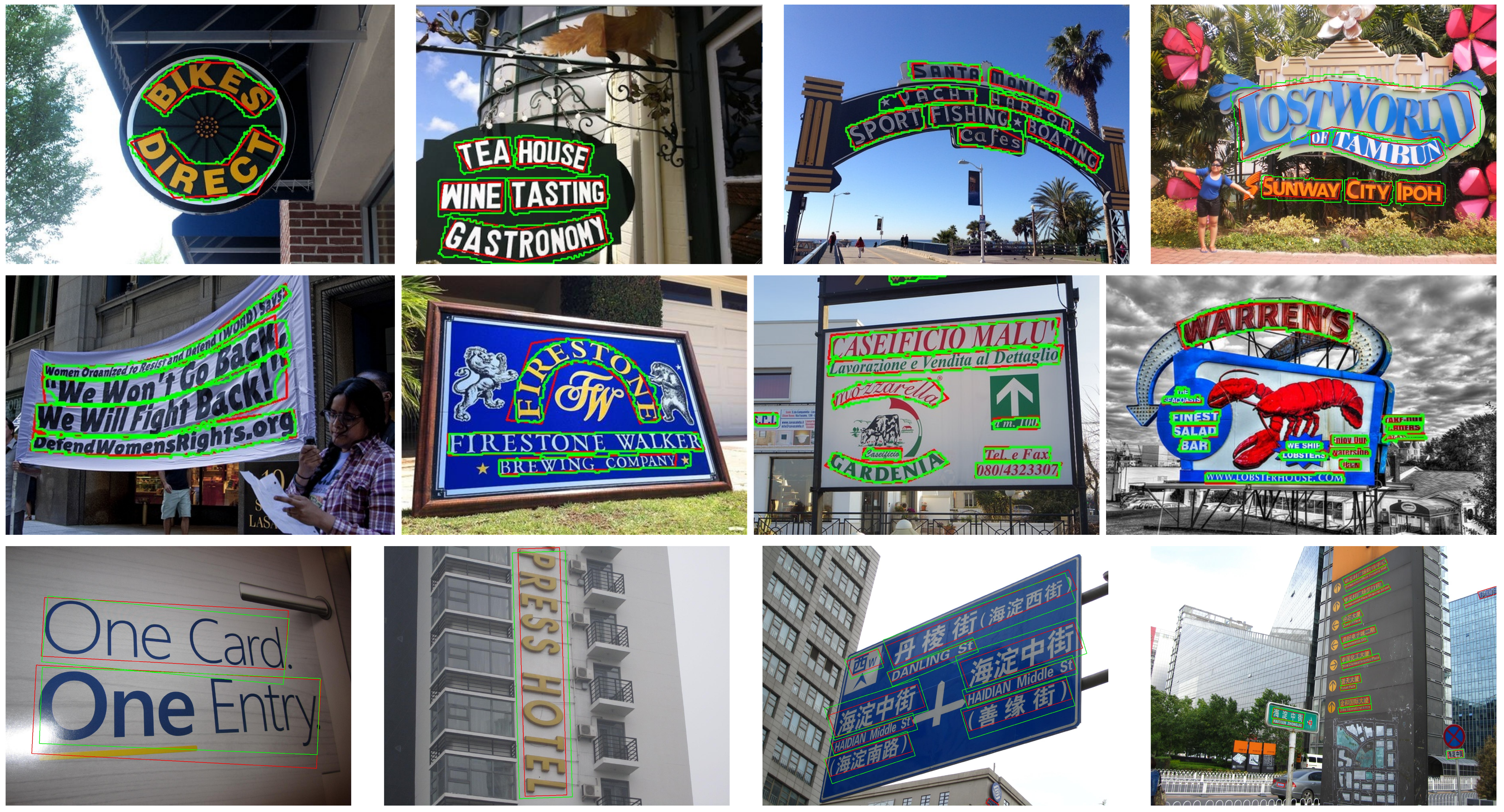}
  \caption{Qualitative results of the proposed method. Images in row 1-3 are
  sampled from Total-Text, CTW1500, and MSRA-TD500, respectively. Ground-truth
  annotations are in red and our detection results are in green.}
\label{fig:instance}
\end{figure*}

We first evaluate our CT on the Total-Text and CTW1500 datasets to examine
its capability for curved text detection. During testing, we set the
short side of images to different scales (320, 512, 640) and keep
their aspect ratios. We compare our methods with other state-of-the-art detectors in
Tab.~\ref{tab:ttctw}. For Total-Text, when learning from scratch, CT-640 achieves
the competitive F-measure of 84.9\%, surpassing most
existing methods pre-trained on external text datasets.
When pre-training on SynthText, the F-measure value of our best model CT-640
reaches 86.3\%, which is 0.6\% better than second-best DRRG~\cite{DRRG},
while still ensuring the real-time detection speed (40.0 FPS).
Fig.~\ref{fig:rank} demonstrates the accuracy-speed trade-off
of some top-performing real-time text detectors, from which it can be observed
that our CT breaks through the limitation of accuracy-speed boundary.
Analogous results can also be obtained on CTW1500. With external training
data, the F-measure of CT-640 is 83.9\%, the second place of all
methods, which is only lower than DGGR. Meanwhile, the speed can
still exceed 40 FPS. In summary, the experiments conducted on
Total-Text and CTW1500 demonstrate that the proposed CT achieves superior or
competitive results compared to state-of-the-art methods,
indicating its superiority in modeling curved texts. We visualize
our detection results in Fig.~\ref{fig:instance} for further inspection.

\paragraph{Multi-oriented text detection}

We also evaluate CT on the MSRA-TD500 dataset to test the robustness
in modeling multi-oriented texts. As shown in Tab.~\ref{tab:msra},
CT achieves the F-measure value of 83.0\% at 34.8 FPS without external
training data. Compared with PAN, our method outperforms it by 4.1\%.
When pre-training on SynthText, the F-measure value of our CT can further be
boosted to 86.1\%. The highest performance and the fastest speed
achieved by CT prove its generalization ability to deal with texts
with extreme aspect ratios and various orientations in complex
natural scenarios.

\begin{wraptable}{l}{7cm}
	\centering
	\caption{Quantitative end-to-end recognition results on Total-Text. The evaluation
	protocol is the same as the one in Mask TextSpotter v3~\cite{masktextspotterv3}.
	“None” means recognition without any lexicon. “Full” lexicon contains all words
	in the test set. * indicates the multi-scale testing is performed.}
\resizebox{7cm}{!}{
	\label{tab:recog}
	\begin{tabular}{l|cc}
		\hline
		Method & None & Full \\
		\hline
		\hline
		TextBoxes*~\cite{textboxes} & 36.3 & 48.9 \\
		Mask TextSpotter v1~\cite{masktextspotterv1} & 52.9 & 71.8 \\
		Qin et al.~\cite{qin} & 63.9 & - \\
		Boundary~\cite{boundary} & 65.0 & 76.1 \\
		Mask TextSpotter v2~\cite{masktextspotterv2} & 65.3 & 77.4 \\
		CharNet*~\cite{charnet} & 69.2 & - \\
		ABCNet*~\cite{abcnet} & 69.5 & 78.4 \\
		Mask TextSpotter v3~\cite{masktextspotterv3} & 71.2 & 78.4 \\
		\textbf{Mask TextSpotter v3 w/ CPN} & \textbf{71.9} & \textbf{79.5} \\
		\hline
	\end{tabular}
}
\end{wraptable}

\paragraph{End-to-end text recognition}

We simply replace SPN in Mask TextSpotter v3 with our proposed CPN
to develop a more powerful end-to-end text recognizer. We evaluate
CPN-based text spotter on Total-Text to test the proposal
generation quality for the text spotting task. As shown in Tab.~\ref{tab:recog},
equipped with CPN, Mask TextSpotter v3 achieves the F-measure values of
71.9\% and 79.5\% when the lexicon is not used and used respectively.
Compared with the original version and other state-of-the-art methods,
our method can obtain higher performance whether the lexicon is provided
or not. Thus, the quantitative results demonstrate that CPN can
produce more accurate text proposals than SPN, which is beneficial for
recognition and can improve the performance of end-to-end text recognition
further.

\begin{figure}[t]
	\centering
	\subfigure[Segmentation Proposal Network (SPN)]{
		\includegraphics[width=0.47\linewidth]{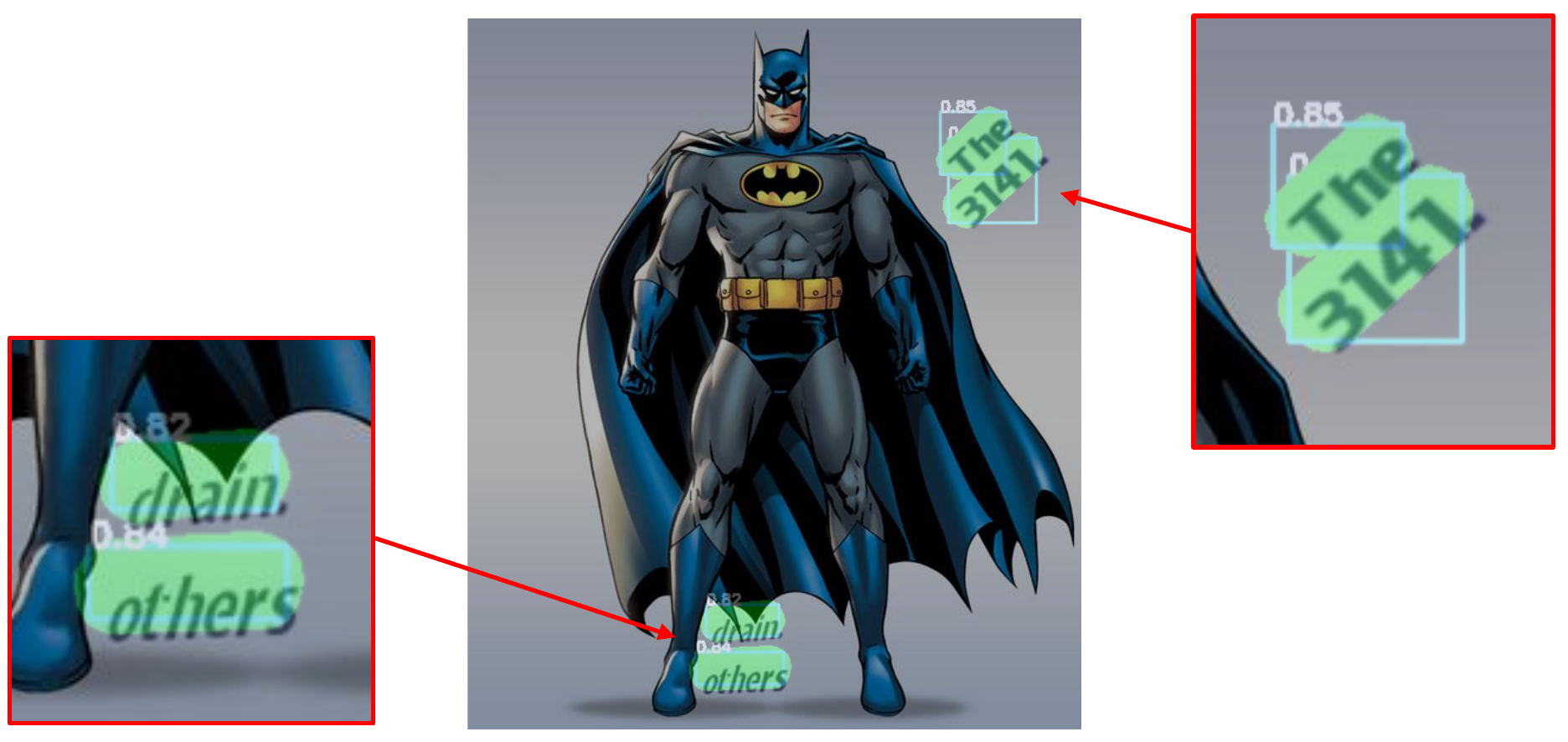}
	}
	\quad
	\subfigure[CentripetalText Proposal Network (CPN)]{
		\includegraphics[width=0.47\linewidth]{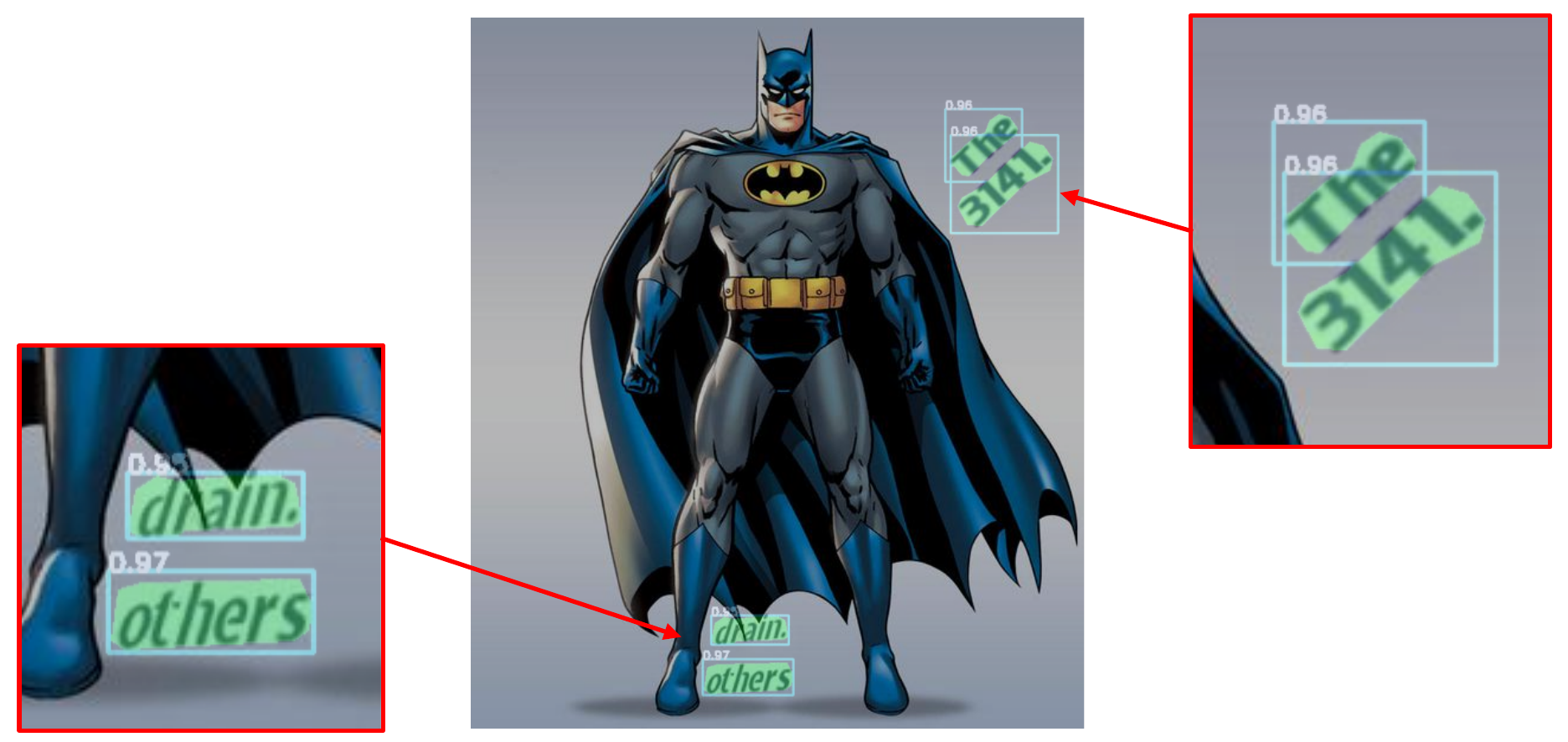}
	}
	\caption{Qualitative comparison of proposals obtained by SPN and CPN. The blue rectangles
	denote the text proposals and the green areas denote the
	binary polygon masks.}
	\label{fig:compare}
\end{figure}

We visualize the text proposals and the polygon masks generated by
SPN and CPN for intuitive comparison. As shown in Fig.~\ref{fig:compare},
we can see that the polygon masks produced by CPN fit the text instances
more tightly, which qualitatively proves the superiority of the proposed
CPN in producing text proposals compared to other approaches.


\section{Conclusion}

To keep the simplicity and robustness of text instance representation, we
proposed CentripetalText (CT) which decomposes text instances into the
combination of text kernels and centripetal shifts. Text kernels
identify the skeletons of text instances while centripetal shifts guide
the external text pixels to the internal text kernels. Moreover, to
reduce the burden of model training, a relaxation operation was integrated
into the dense regression for centripetal shifts, allowing
the correct prediction in a range. Equipped with the proposed CT, our detector
achieved superior or comparable performance compared to other state-of-the-art
methods while keeping the real-time inference speed. The source code is
available at \url{https://github.com/shengtao96/CentripetalText}, and we
hope that the proposed CT can serve as a valuable and common representation
for scene texts.


\begin{ack}

This work was supported by Beijing Nova Program of Science and Technology (Grant No.: Z191100001119077), Project 2020BD020 supported by PKU-Baidu Fund, National Language Committee of China (Grant No.: ZDI135-130), Center For Chinese Font Design and Research, Key Laboratory of Science, Technology and Standard in Press Industry (Key Laboratory of Intelligent Press Media Technology), State Key Laboratory of Media Convergence Production Technology and Systems.
\end{ack}

{
\small
\bibliographystyle{plain}
\bibliography{mybibliography}

\begin{thebibliography}{10}

\bibitem{CRAFT}
Youngmin Baek, Bado Lee, Dongyoon Han, Sangdoo Yun, and Hwalsuk Lee.
\newblock Character region awareness for text detection.
\newblock In {\em CVPR}, pages 9365--9374, 2019.

\bibitem{totaltext}
Chee~Kheng Ch'ng and Chee~Seng Chan.
\newblock Total-text: A comprehensive dataset for scene text detection and
  recognition.
\newblock In {\em ICDAR}, volume~1, pages 935--942. IEEE, 2017.

\bibitem{pixellink}
Dan Deng, Haifeng Liu, Xuelong Li, and Deng Cai.
\newblock Pixellink: Detecting scene text via instance segmentation.
\newblock In {\em AAAI}, 2018.

\bibitem{imagenet}
Jia Deng, Wei Dong, Richard Socher, Li-Jia Li, Kai Li, and Li~Fei-Fei.
\newblock Imagenet: A large-scale hierarchical image database.
\newblock In {\em CVPR}, pages 248--255. Ieee, 2009.

\bibitem{fastrcnn}
Ross Girshick.
\newblock Fast r-cnn.
\newblock In {\em ICCV}, pages 1440--1448, 2015.

\bibitem{synthtext}
Ankush Gupta, Andrea Vedaldi, and Andrew Zisserman.
\newblock Synthetic data for text localisation in natural images.
\newblock In {\em CVPR}, pages 2315--2324, 2016.

\bibitem{maskrcnn}
Kaiming He, Georgia Gkioxari, Piotr Doll{\'a}r, and Ross Girshick.
\newblock Mask r-cnn.
\newblock In {\em ICCV}, pages 2961--2969, 2017.

\bibitem{resnet}
Kaiming He, Xiangyu Zhang, Shaoqing Ren, and Jian Sun.
\newblock Deep residual learning for image recognition.
\newblock In {\em CVPR}, pages 770--778, 2016.

\bibitem{icdar2015}
Dimosthenis Karatzas, Lluis Gomez-Bigorda, Anguelos Nicolaou, Suman Ghosh,
  Andrew Bagdanov, Masakazu Iwamura, Jiri Matas, Lukas Neumann,
  Vijay~Ramaseshan Chandrasekhar, Shijian Lu, et~al.
\newblock Icdar 2015 competition on robust reading.
\newblock In {\em ICDAR}, pages 1156--1160. IEEE, 2015.

\bibitem{icdar2013}
Dimosthenis Karatzas, Faisal Shafait, Seiichi Uchida, Masakazu Iwamura,
  Lluis~Gomez i~Bigorda, Sergi~Robles Mestre, Joan Mas, David~Fernandez Mota,
  Jon~Almazan Almazan, and Lluis~Pere De~Las~Heras.
\newblock Icdar 2013 robust reading competition.
\newblock In {\em ICDAR}, pages 1484--1493. IEEE, 2013.

\bibitem{GFL}
Xiang Li, Wenhai Wang, Lijun Wu, Shuo Chen, Xiaolin Hu, Jun Li, Jinhui Tang,
  and Jian Yang.
\newblock Generalized focal loss: Learning qualified and distributed bounding
  boxes for dense object detection.
\newblock In {\em NIPS}, volume~33, pages 21002--21012. Curran Associates,
  Inc., 2020.

\bibitem{masktextspotterv2}
Minghui Liao, Pengyuan Lyu, Minghang He, Cong Yao, Wenhao Wu, and Xiang Bai.
\newblock Mask textspotter: An end-to-end trainable neural network for spotting
  text with arbitrary shapes.
\newblock {\em IEEE Transactions on Pattern Analysis and Machine Intelligence},
  2019.

\bibitem{masktextspotterv3}
Minghui Liao, Guan Pang, Jing Huang, Tal Hassner, and Xiang Bai.
\newblock Mask textspotter v3: Segmentation proposal network for robust scene
  text spotting.
\newblock In {\em ECCV}, pages 706--722, Cham, 2020. Springer International
  Publishing.

\bibitem{textboxes++}
Minghui Liao, Baoguang Shi, and Xiang Bai.
\newblock Textboxes++: A single-shot oriented scene text detector.
\newblock {\em IEEE Trans. Image Process.}, 27(8):3676--3690, 2018.

\bibitem{textboxes}
Minghui Liao, Baoguang Shi, Xiang Bai, Xinggang Wang, and Wenyu Liu.
\newblock Textboxes: A fast text detector with a single deep neural network.
\newblock In {\em AAAI}, volume~31, 2017.

\bibitem{DB}
Minghui Liao, Zhaoyi Wan, Cong Yao, Kai Chen, and Xiang Bai.
\newblock Real-time scene text detection with differentiable binarization.
\newblock In {\em AAAI}, 2020.

\bibitem{rrd}
Minghui Liao, Zhen Zhu, Baoguang Shi, Gui-song Xia, and Xiang Bai.
\newblock Rotation-sensitive regression for oriented scene text detection.
\newblock In {\em CVPR}, pages 5909--5918, 2018.

\bibitem{fpn}
Tsung-Yi Lin, Piotr Doll{\'a}r, Ross Girshick, Kaiming He, Bharath Hariharan,
  and Serge Belongie.
\newblock Feature pyramid networks for object detection.
\newblock In {\em CVPR}, pages 2117--2125, 2017.

\bibitem{ssd}
Wei Liu, Dragomir Anguelov, Dumitru Erhan, Christian Szegedy, Scott Reed,
  Cheng-Yang Fu, and Alexander~C Berg.
\newblock Ssd: Single shot multibox detector.
\newblock In {\em ECCV}, pages 21--37. Springer, 2016.

\bibitem{abcnet}
Yuliang Liu, Hao Chen, Chunhua Shen, Tong He, Lianwen Jin, and Liangwei Wang.
\newblock Abcnet: Real-time scene text spotting with adaptive bezier-curve
  network.
\newblock In {\em CVPR}, pages 9809--9818, 2020.

\bibitem{MCN}
Zichuan Liu, Guosheng Lin, Sheng Yang, Jiashi Feng, Weisi Lin, and Wang~Ling
  Goh.
\newblock Learning markov clustering networks for scene text detection.
\newblock In {\em CVPR}, pages 6936--6944. IEEE, 2018.

\bibitem{fcn}
Jonathan Long, Evan Shelhamer, and Trevor Darrell.
\newblock Fully convolutional networks for semantic segmentation.
\newblock In {\em CVPR}, pages 3431--3440, 2015.

\bibitem{textsnake}
Shangbang Long, Jiaqiang Ruan, Wenjie Zhang, Xin He, Wenhao Wu, and Cong Yao.
\newblock Textsnake: A flexible representation for detecting text of arbitrary
  shapes.
\newblock In {\em ECCV}, pages 20--36, 2018.

\bibitem{masktextspotterv1}
Pengyuan Lyu, Minghui Liao, Cong Yao, Wenhao Wu, and Xiang Bai.
\newblock Mask textspotter: An end-to-end trainable neural network for spotting
  text with arbitrary shapes.
\newblock In {\em ECCV}, pages 67--83, 2018.

\bibitem{rrpn}
Jianqi Ma, Weiyuan Shao, Hao Ye, Li~Wang, Hong Wang, Yingbin Zheng, and
  Xiangyang Xue.
\newblock Arbitrary-oriented scene text detection via rotation proposals.
\newblock {\em IEEE Transactions on Multimedia}, 20(11):3111--3122, 2018.

\bibitem{dice}
Fausto Milletari, Nassir Navab, and Seyed-Ahmad Ahmadi.
\newblock V-net: Fully convolutional neural networks for volumetric medical
  image segmentation.
\newblock In {\em 3DV}, pages 565--571. IEEE, 2016.

\bibitem{librarcnn}
Jiangmiao Pang, Kai Chen, Jianping Shi, Huajun Feng, Wanli Ouyang, and Dahua
  Lin.
\newblock Libra r-cnn: Towards balanced learning for object detection.
\newblock In {\em CVPR}, pages 821--830, 2019.

\bibitem{qin}
Siyang Qin, Alessandro Bissacco, Michalis Raptis, Yasuhisa Fujii, and Ying
  Xiao.
\newblock Towards unconstrained end-to-end text spotting.
\newblock In {\em ICCV}, pages 4704--4714, 2019.

\bibitem{fasterrcnn}
Shaoqing Ren, Kaiming He, Ross Girshick, and Jian Sun.
\newblock Faster r-cnn: Towards real-time object detection with region proposal
  networks.
\newblock In {\em NIPS}, pages 91--99, 2015.

\bibitem{BR}
Tao Sheng and Zhouhui Lian.
\newblock Bidirectional regression for arbitrary-shaped text detection.
\newblock In {\em Document Analysis and Recognition -- ICDAR 2021}, pages
  187--201, 2021.

\bibitem{seglink}
Baoguang Shi, Xiang Bai, and Serge Belongie.
\newblock Detecting oriented text in natural images by linking segments.
\newblock In {\em CVPR}, pages 2550--2558, 2017.

\bibitem{ohem}
Abhinav Shrivastava, Abhinav Gupta, and Ross Girshick.
\newblock Training region-based object detectors with online hard example
  mining.
\newblock In {\em CVPR}, pages 761--769, 2016.

\bibitem{seglink++}
Jun Tang, Zhibo Yang, Yongpan Wang, Qi~Zheng, Yongchao Xu, and Xiang Bai.
\newblock Seglink++: Detecting dense and arbitrary-shaped scene text by
  instance-aware component grouping.
\newblock {\em PR}, 96:106954, 2019.

\bibitem{CTPN}
Zhi Tian, Weilin Huang, Tong He, Pan He, and Yu~Qiao.
\newblock Detecting text in natural image with connectionist text proposal
  network.
\newblock In {\em ECCV}, pages 56--72. Springer, 2016.

\bibitem{vatti}
Bala~R Vatti.
\newblock A generic solution to polygon clipping.
\newblock {\em Communications of the ACM}, 35(7):56--63, 1992.

\bibitem{boundary}
Hao Wang, Pu~Lu, Hui Zhang, Mingkun Yang, Xiang Bai, Yongchao Xu, Mengchao He,
  Yongpan Wang, and Wenyu Liu.
\newblock All you need is boundary: Toward arbitrary-shaped text spotting.
\newblock In {\em AAAI}, volume~34, pages 12160--12167, 2020.

\bibitem{psenet}
Wenhai Wang, Enze Xie, Xiang Li, Wenbo Hou, Tong Lu, Gang Yu, and Shuai Shao.
\newblock Shape robust text detection with progressive scale expansion network.
\newblock In {\em CVPR}, pages 9336--9345, 2019.

\bibitem{PAN}
Wenhai Wang, Enze Xie, Xiaoge Song, Yuhang Zang, Wenjia Wang, Tong Lu, Gang Yu,
  and Chunhua Shen.
\newblock Efficient and accurate arbitrary-shaped text detection with pixel
  aggregation network.
\newblock In {\em ICCV}, pages 8440--8449, 2019.

\bibitem{contournet}
Yuxin Wang, Hongtao Xie, Zheng-Jun Zha, Mengting Xing, Zilong Fu, and Yongdong
  Zhang.
\newblock Contournet: Taking a further step toward accurate arbitrary-shaped
  scene text detection.
\newblock In {\em CVPR}, pages 11753--11762, 2020.

\bibitem{spcnet}
Enze Xie, Yuhang Zang, Shuai Shao, Gang Yu, Cong Yao, and Guangyao Li.
\newblock Scene text detection with supervised pyramid context network.
\newblock In {\em AAAI}, volume~33, pages 9038--9045, 2019.

\bibitem{charnet}
Linjie Xing, Zhi Tian, Weilin Huang, and Matthew~R Scott.
\newblock Convolutional character networks.
\newblock In {\em ICCV}, pages 9126--9136, 2019.

\bibitem{textfield}
Yongchao Xu, Yukang Wang, Wei Zhou, Yongpan Wang, Zhibo Yang, and Xiang Bai.
\newblock Textfield: Learning a deep direction field for irregular scene text
  detection.
\newblock {\em IEEE Transactions on Image Processing}, 28(11):5566--5579, 2019.

\bibitem{MSR}
Chuhui Xue, Shijian Lu, and Wei Zhang.
\newblock Msr: multi-scale shape regression for scene text detection.
\newblock In {\em IJCAI}, pages 989--995. AAAI Press, 2019.

\bibitem{hust}
Cong Yao, Xiang Bai, and Wenyu Liu.
\newblock A unified framework for multioriented text detection and recognition.
\newblock {\em IEEE Transactions on Image Processing}, 23(11):4737--4749, 2014.

\bibitem{msratd500}
Cong Yao, Xiang Bai, Wenyu Liu, Yi~Ma, and Zhuowen Tu.
\newblock Detecting texts of arbitrary orientations in natural images.
\newblock In {\em CVPR}, pages 1083--1090. IEEE, 2012.

\bibitem{bisenet}
Changqian Yu, Jingbo Wang, Chao Peng, Changxin Gao, Gang Yu, and Nong Sang.
\newblock Bisenet: Bilateral segmentation network for real-time semantic
  segmentation.
\newblock In {\em ECCV}, pages 325--341, 2018.

\bibitem{ctw1500}
Liu Yuliang, Jin Lianwen, Zhang Shuaitao, and Zhang Sheng.
\newblock Detecting curve text in the wild: New dataset and new solution.
\newblock {\em arXiv preprint arXiv:1712.02170}, 2017.

\bibitem{LOMO}
Chengquan Zhang, Borong Liang, Zuming Huang, Mengyi En, Junyu Han, Errui Ding,
  and Xinghao Ding.
\newblock Look more than once: An accurate detector for text of arbitrary
  shapes.
\newblock In {\em CVPR}, pages 10552--10561, 2019.

\bibitem{DRRG}
Shi-Xue Zhang, Xiaobin Zhu, Jie-Bo Hou, Chang Liu, Chun Yang, Hongfa Wang, and
  Xu-Cheng Yin.
\newblock Deep relational reasoning graph network for arbitrary shape text
  detection.
\newblock In {\em CVPR}, pages 9699--9708, 2020.

\bibitem{zhaoetal}
Hengshuang Zhao, Jianping Shi, Xiaojuan Qi, Xiaogang Wang, and Jiaya Jia.
\newblock Pyramid scene parsing network.
\newblock In {\em CVPR}, pages 2881--2890, 2017.

\bibitem{east}
Xinyu Zhou, Cong Yao, He~Wen, Yuzhi Wang, Shuchang Zhou, Weiran He, and Jiajun
  Liang.
\newblock East: an efficient and accurate scene text detector.
\newblock In {\em CVPR}, pages 5551--5560, 2017.

\end{thebibliography}
}

\newpage

\appendix

\section{Rotation robustness analysis}

\begin{table}[H]
	\caption{Quantitative end-to-end recognition results (without lexicon) on Rotated
	ICDAR2013. The evaluation protocol is the same as the one in ICDAR2015 dataset.
	CharNet is tested with the official released model. Mask TextSpotter v2 (MTSv2),
	Mask TextSpotter v3 (MTSv3) and our model (MTSv3 w/ CPN) are trained with the same
	rotating augmentation. “RA” is short for rotating angles. “P”, “R” and “F” represent
	the precision, recall and F-measure respectively.}
	\label{tab:ro13}
	\begin{tabular}{c|ccc|ccc|ccc|ccc}
		\hline
		\multirow{2}{*}{RA($^{\circ}$)} & \multicolumn{3}{c|}{CharNet~\cite{charnet}} & \multicolumn{3}{c|}{MTSv2~\cite{masktextspotterv2}} & \multicolumn{3}{c|}{MTSv3~\cite{masktextspotterv3}} & \multicolumn{3}{c}{\textbf{MTSv3 w/ CPN}} \\
		\cline{2-13}
		 & P & R & F & P & R & F & P & R & F & P & R & F \\
		\hline
		\hline
		0 & 61.7 & 61.2 & 61.4 & 86.3 & 75.2 & 80.3 & 89.0 & 73.0 & 80.2 & 89.7 & 76.3 & 82.4 \\
		15 & 66.3 & 61.9 & 64.0 & 78.4 & 53.5 & 63.6 & 87.2 & 69.8 & 77.5 & 87.8 & 72.0 & 79.1 \\
		30 & 60.9 & 56.5 & 58.6 & 73.9 & 54.7 & 62.9 & 87.8 & 67.5 & 76.3 & 89.6 & 69.4 & 78.2 \\
		45 & 34.2 & 33.5 & 33.9 & 66.4 & 45.8 & 54.2 & 88.5 & 66.8 & 76.1 & 89.4 & 66.9 & 76.5 \\
		60 & 10.3 & 8.4 & 9.3 & 68.2 & 48.3 & 56.6 & 88.5 & 67.6 & 76.6 & 88.6 & 67.1 & 76.4 \\
		75 & 0.3 & 0.2 & 0.2 & 77.0 & 59.2 & 67.0 & 86.9 & 67.6 & 76.0 & 88.1 & 67.7 & 76.5 \\
		90 & 0.0 & 0.0 & 0.0 & 82.0 & 56.9 & 67.1 & 85.9 & 57.9 & 69.1 & 87.8 & 60.6 & 71.7 \\
		\hline
	\end{tabular}
\end{table}

To further demonstrate the rotation robustness of our method, we evaluate
our CPN-based text spotter on the Rotated ICDAR2013 dataset.

\textbf{Rotated ICDAR2013}~\cite{masktextspotterv3} is an augmented text
dataset that is generated from ICDAR2013~\cite{icdar2013}.
To form the Rotated ICDAR2013 dataset, all the images and annotations
in the test set of the ICDAR2013 benchmark are rotated with some specific
angles, including $15^{\circ}$, $30^{\circ}$, $45^{\circ}$, $60^{\circ}$,
$75^{\circ}$ and $90^{\circ}$. The dataset contains 229 training images
and 233 testing images. The text instances are annotated at the
text-line level with rotated rectangles. Since the annotations are extended
from horizontal rectangles to multi-oriented ones, we adopt the evaluation
protocols in the ICDAR2015 dataset~\cite{icdar2015}.

As shown in Tab.~\ref{tab:ro13}, we compare three top-performing methods
CharNet~\cite{charnet}, Mask TextSpotter v2~\cite{masktextspotterv2}, and
Mask TextSpotter v3~\cite{masktextspotterv3} with our proposed text spotter
at different rotation angles. We can see that CharNet and Mask TextSpotter
v2 fail to deal with the multi-oriented texts and their performances fall well
below ours. Moreover, Our method surpasses Mask TextSpotter v3 by more than 1.5\%
when the rotation angles are $0^{\circ}$, $15^{\circ}$, $30^{\circ}$ and
$90^{\circ}$, and we obtain the competitive performance under the other
angles. The extensive experiments prove the superior robustness to
various orientations of scene texts offered by our method.

\section{Limitation}

\begin{figure}[h]
	\centering
	\subfigure[]{
		\includegraphics[width=0.47\linewidth]{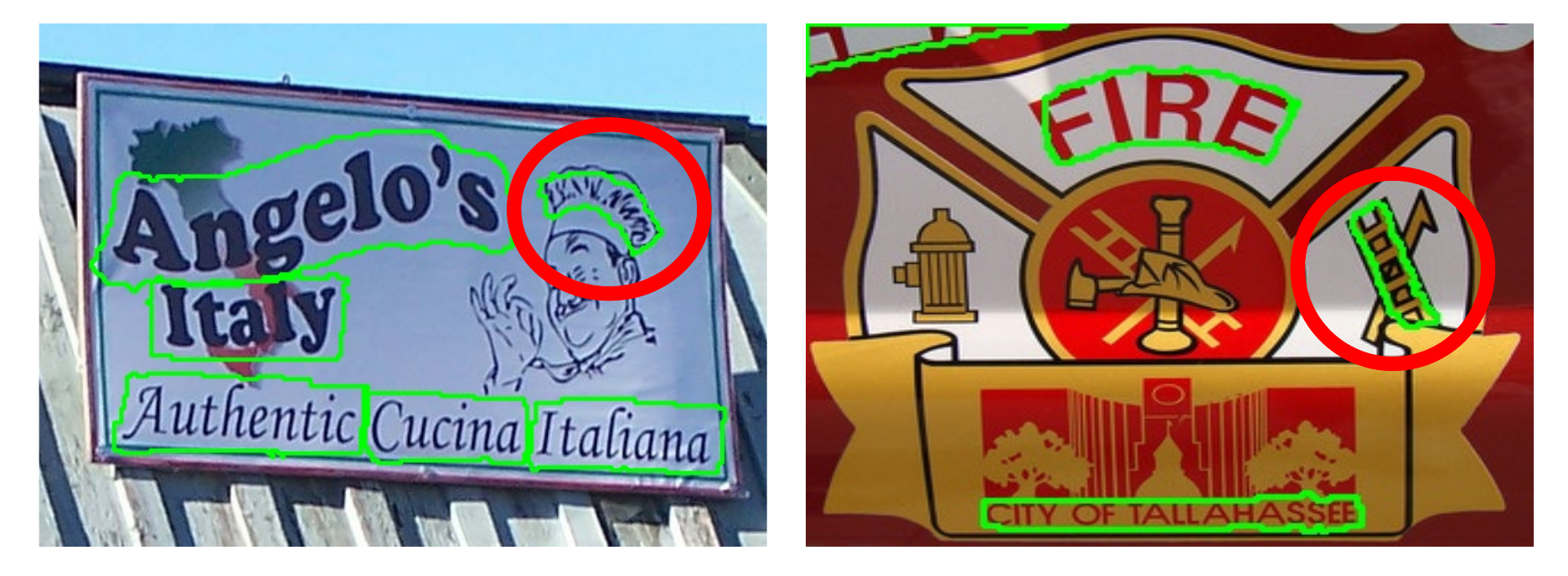}
		\label{fig:failexamplesA}
	}
	\quad
	\subfigure[]{
		\includegraphics[width=0.47\linewidth]{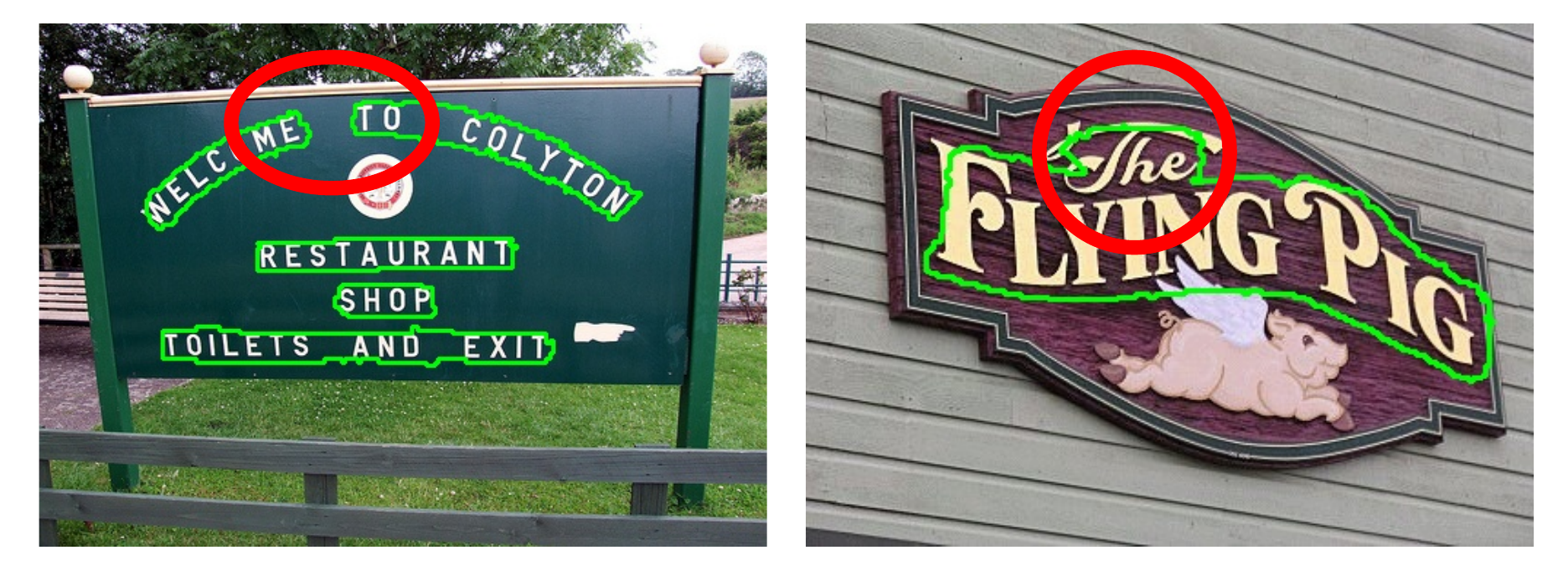}
		\label{fig:failexamplesB}
	}
	\caption{Failure samples.}
	\label{fig:failexamples}
\end{figure}

Although the proposal CT works well in most cases of scene text
detection, it still fails in some difficult cases as shown in
Fig.~\ref{fig:failexamples}. On the one hand, our method may mistakenly treat
some decorative patterns as texts and thus produces false
positives (see Fig.~\ref{fig:failexamplesA}). In this situation, the
followed recognition module can effectively restrain such failures
according to the high-level semantic information. On the other hand,
whether two close text instance should be connected into one or
not is still a challenging problem which influences the detection
performance deeply (see Fig.~\ref{fig:failexamplesB}). In the future,
we plan to solve this problem and make the model more robust.

\section{More detection and recognition results}

\begin{figure}[H]
  \centering
  \includegraphics[width=\linewidth]{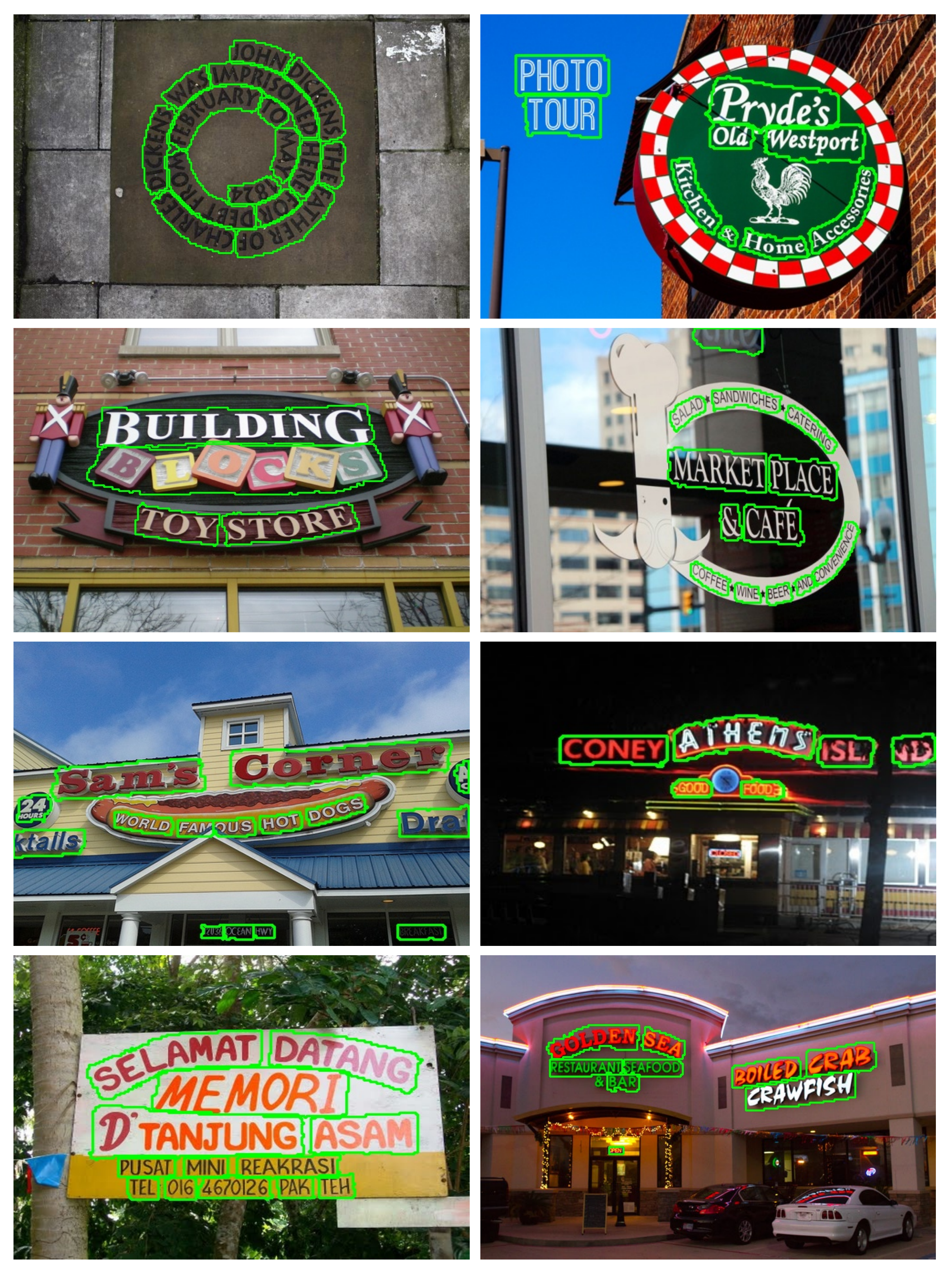}
  \caption{Detection results on Total-Text.}
\label{fig:mtt}
\end{figure}

\begin{figure}[H]
  \centering
  \includegraphics[width=\linewidth]{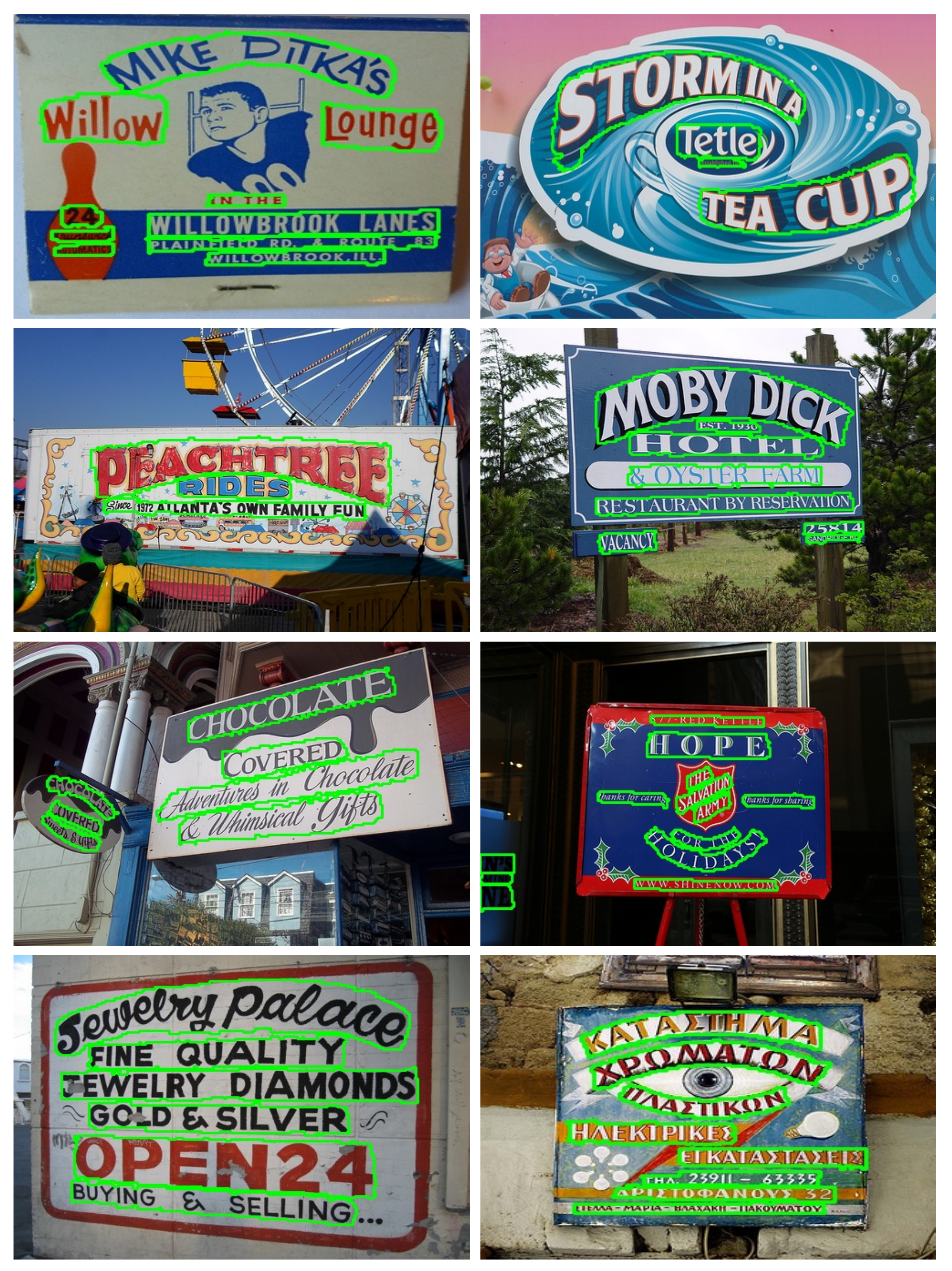}
  \caption{Detection results on CTW1500.}
\label{fig:mctw}
\end{figure}

\begin{figure}[H]
  \centering
  \includegraphics[width=\linewidth]{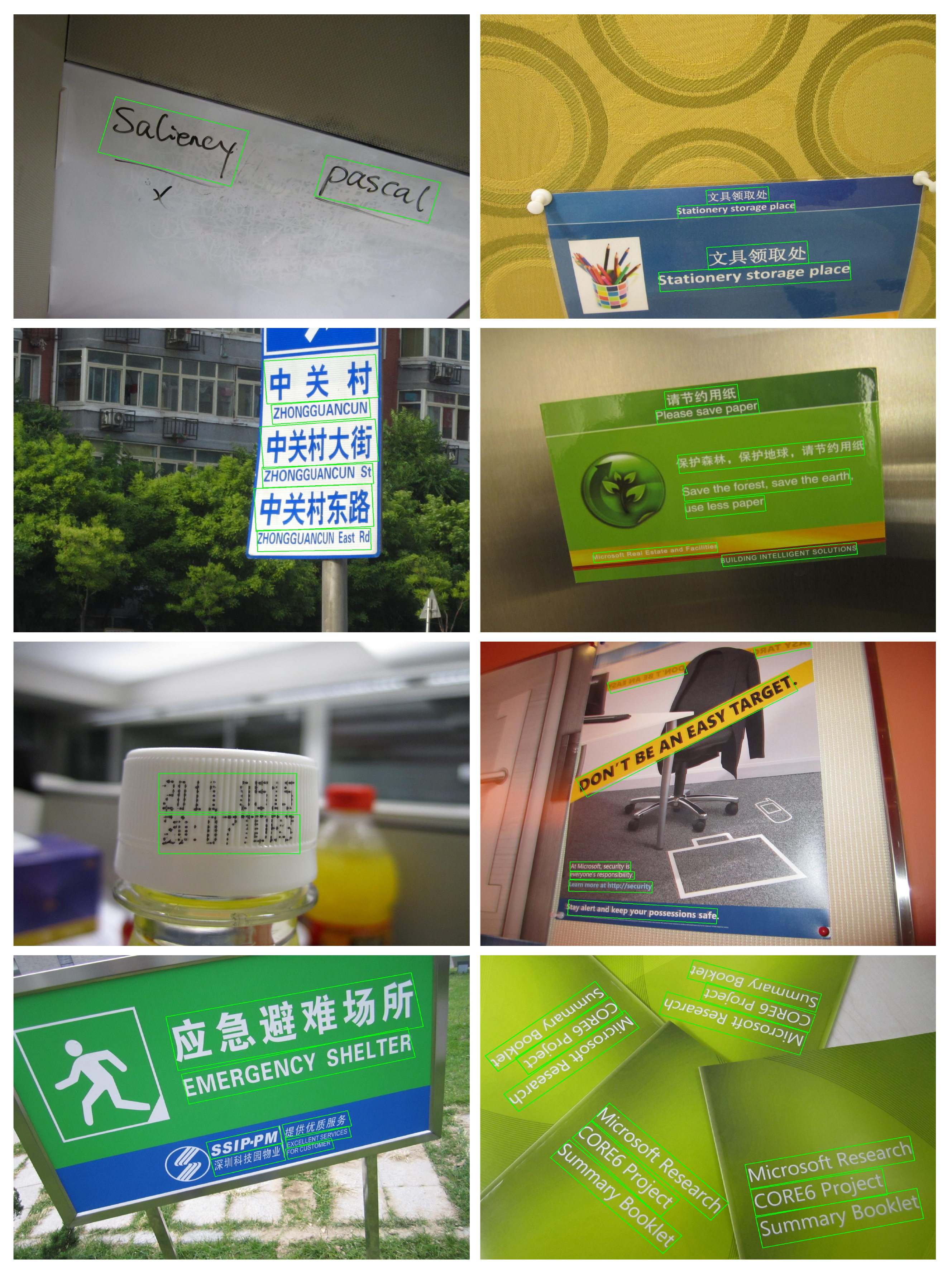}
  \caption{Detection results on MSRA-TD500.}
\label{fig:mmsra}
\end{figure}

\begin{figure}[H]
  \centering
  \includegraphics[width=\linewidth]{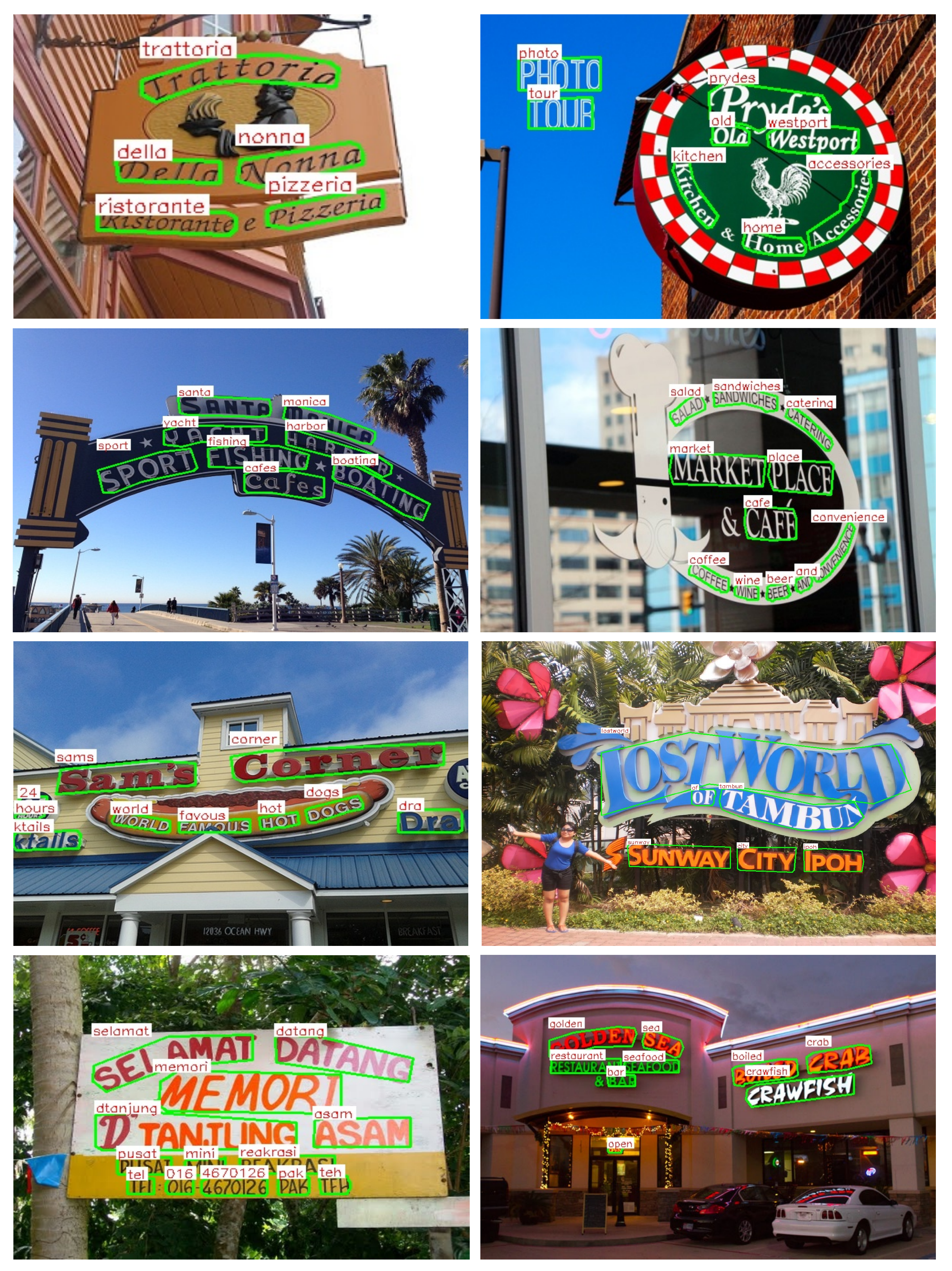}
  \caption{Recognition results on Total-Text.}
\label{fig:mttreg}
\end{figure}

More detection results are shown in Fig.~\ref{fig:mtt} (Total-Text),
Fig.~\ref{fig:mctw} (CTW1500), and Fig.~\ref{fig:mmsra} (MSRA-TD500),
and end-to-end recognition results on Total-Text are shown in Fig.~\ref{fig:mttreg}.

\end{document}